\documentclass[acmsmall,authorversion,nonacm]{acmart} % remove the copyright space and footnote
\usepackage{graphicx}
\usepackage{subfigure}
\usepackage{amsfonts}
\usepackage{wrapfig}

\usepackage[ruled,linesnumbered]{algorithm2e}

\usepackage{array}
\usepackage{booktabs}
\usepackage{multirow}
\usepackage{amsmath}
\AtBeginDocument{%
  }

\setcopyright{acmcopyright}
\copyrightyear{2018}
\acmYear{2018}
\acmDOI{XXXXXXX.XXXXXXX}

\acmJournal{JACM}
\acmVolume{37}
\acmNumber{4}
\acmArticle{111}
\acmMonth{8}

\begin{document}

\title{Saliency Attack: Towards Imperceptible Black-box Adversarial Attack}

\author{Zeyu Dai}
\orcid{0000-0002-1351-476X}
\affiliation{%
  \institution{Southern University of Science and Technology}
  % \streetaddress{1088 Xueyuan Avenue}
  % \city{Shenzhen}
  % \country{China}
}
\affiliation{%
  \institution{The Hong Kong Polytechnic University}
  % \streetaddress{11 Yuk Choi Road, Hung Hom, Kowloon}
  % \city{Hong Kong}
  \country{China}
}
\email{daizy2019@mail.sustech.edu.cn}

\author{Shengcai Liu}
\authornote{Corresponding Author}
\affiliation{%
  \institution{Southern University of Science and Technology}
  % \city{Shenzhen}
  \country{China}
}
\email{liusc3@sustech.edu.cn}

\author{Ke Tang}
\affiliation{%
  \institution{Southern University of Science and Technology}
  % \city{Shenzhen}
  \country{China}
}
\email{tangk3@sustech.edu.cn}

\author{Qing Li}
\affiliation{%
  \institution{The Hong Kong Polytechnic University}
  % \city{Hong Kong}
  \country{China}
}
\email{qing-prof.li@polyu.edu.hk}


\begin{abstract}
  Deep neural networks are vulnerable to adversarial examples, even in the black-box setting where the attacker is only accessible to the model output.
  Recent studies have devised effective black-box attacks with high query efficiency.
  However, such performance is often accompanied by compromises in attack imperceptibility, hindering the practical use of these approaches.
  In this paper, we propose to restrict the perturbations to a small \textit{salient} region to generate adversarial examples that can hardly be perceived.
  This approach is readily compatible with many existing black-box attacks and can significantly improve their imperceptibility with little degradation in attack success rate.
  Further, we propose the Saliency Attack, a new black-box attack aiming to refine the perturbations in the salient region to achieve even better imperceptibility.
  Extensive experiments show that compared to the state-of-the-art black-box attacks, our approach achieves much better imperceptibility scores, including most apparent distortion (MAD), $L_0$ and $L_2$ distances, and also obtains significantly higher success rates judged by a human-like threshold on MAD.
  Importantly, the perturbations generated by our approach are interpretable to some extent.
  Finally, it is also demonstrated to be robust to different detection-based defenses.
\end{abstract}


%    <concept>
%        <concept_id>10002978</concept_id>
%        <concept_desc>Security and privacy</concept_desc>
%        <concept_significance>500</concept_significance>
%        </concept>
%    <concept>
%        <concept_id>10010147.10010178.10010224</concept_id>
%        <concept_desc>Computing methodologies~Computer vision</concept_desc>
%        <concept_significance>500</concept_significance>
%        </concept>
%  </ccs2012>
% \ccsdesc[500]{Security and privacy}
% \ccsdesc[500]{Computing methodologies~Computer vision}


% \keywords{Deep neural networks, black-box adversarial attack, salient object detection}

\maketitle

\section{Introduction}
Deep neural networks (DNNs) have achieved significant progress in wide applications, such as image classification \cite{1}, face recognition \cite{2}, object detection \cite{3}, speech recognition \cite{4} and machine translation \cite{5}. Despite their success, deep learning models have exhibited vulnerability to adversarial attacks \cite{6,7,53,54}. Crafted by adding some small perturbations to benign inputs, adversarial examples (AEs) can fool DNNs into making wrong predictions, which is a critical threat especially for some security-sensitive scenarios such as autonomous driving \cite{8}.

Existing adversarial attacks can be divided into \textit{white-box} and \textit{black-box} attacks according to the accessibility to the target model. White-box attacks \cite{6,7,9,10,11} have full access to the architecture and parameters of the target model, and can generate successful AEs easily via backpropagation. However, in practice, the model internals are often unavailable to attackers. This gives rise to the more realistic and challenging black-box attacks \cite{12,13,14,15,16,17} that only require the output of the target model.

\begin{figure}
	\centering
	\subfigure[Original image]{
		\begin{minipage}[b]{0.22\columnwidth}
			\includegraphics[width=\columnwidth]{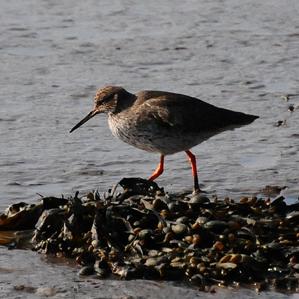}
	\end{minipage}}%
	\subfigure[Parsimonious Attack]{
		\begin{minipage}[b]{0.22\columnwidth}
			\includegraphics[width=\columnwidth]{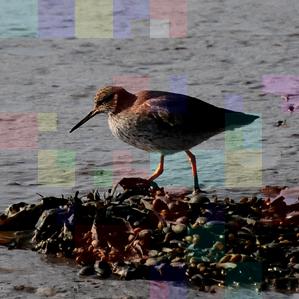}
	\end{minipage}}%
	\subfigure[Square Attack]{
		\begin{minipage}[b]{0.22\columnwidth}
			\includegraphics[width=\columnwidth]{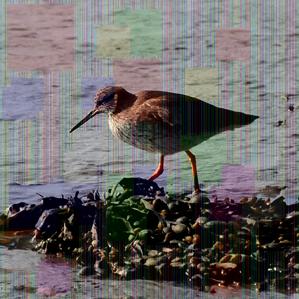}
	\end{minipage}}%
	% \vspace{-1\baselineskip}
	\caption{Some AEs generated by the SOTA black-box attacks.}\label{fig2}

\end{figure}

Motivated by the fact that many real-world online application programming interfaces (APIs) often pose mandatory time or monetary limits to user queries \cite{13}, most recent research on black-box attacks concerns improving query efficiency, which indeed has achieved notable progress. For example, the state-of-the-art (SOTA) Square Attack \cite{16} can succeed in untargeted attack on ImageNet dataset \cite{1} with only tens of queries on average. On the other hand, such performance is often achieved by applying large-region or even global perturbations (e.g., random vertical stripes in Square Attack) to the original input, eventually resulting in unnatural AEs with significant visual differences from the original (see Figure~\ref{fig2} for some examples).
In effect, imperceptibility is essential for attackers. Current online APIs usually integrate detectors into their services to detect anomalous inputs \cite{17}. Those AEs with too visible perturbation are difficult to pass these detectors, not to mention human judgment, hindering the practical use of these black-box attacks.

\begin{figure}[tbp]
	\setlength{\belowcaptionskip}{-6pt}
	\centering
	\scalebox{0.9}{
	\includegraphics[width=\columnwidth]{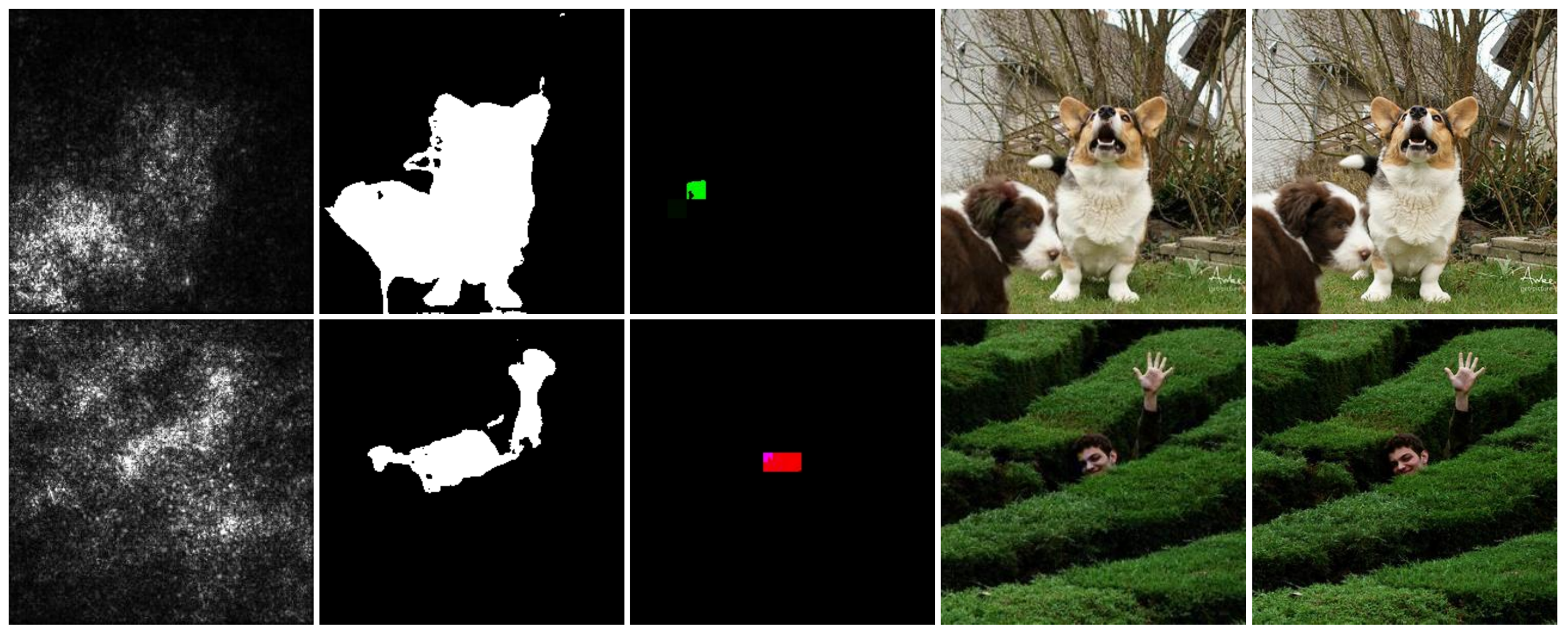}}
	\caption{An illustration of our Saliency Attack. \textbf{1st column:} BP-saliency map, specific to a given image and the corresponding class;  \textbf{2nd column:} salient region, generated by salient object detection and roughly accords with the region of salient pixels in BP-saliency map; \textbf{3rd column:} perturbation, generated by Saliency Attack and restricted in salient regions; \textbf{4th column:} final adversarial example; \textbf{5th column:} original image.}\label{fig1}
\end{figure}

This paper aims to address the above issue.
The key insight is that we can lead black-box attacks to search in a small \textit{salient} region that has the most significant influences on the model output, to achieve successful attacks with imperceptible perturbations.
Concretely, it has been shown in previous DNNs visualization work \cite{18} that one can generate the so-called BP-saliency map to illustrate which features can influence the model prediction by calculating derivatives of the model output with respect to the input.
As shown in the first column of Figure~\ref{fig1}, the brighter pixels in the BP-saliency map have greater impacts on the model output, and therefore perturbing them is more likely to alter the model prediction.
Actually, the classical white-box attack JSMA \cite{9} constructs exactly such a map and iteratively selects pixels from it to perturb.
Despite the fact that the BP-saliency map cannot be used for black-box attacks due to inaccessibility to model gradients, one can observe from the second column of Figure~\ref{fig1} that the region of bright pixels roughly represents the position of the main object in the image.
That is to say, in black-box settings, we can leverage the existing salient object detection model \cite{23} that requires no information other than the input image to approximately obtain the salient region, and then restrict the perturbations to it.
This approach is appealing because it is readily compatible with most existing black-box attacks.
By integrating it into SOTA black-box attacks, we find the modified attacks enjoy significant improvement in attack imperceptibility, with little degradation in success rate (see the results in Table~\ref{tab2}).
Nevertheless, they still suffer from the strategy of perturbing globally, eventually generating complex and irregular perturbations that span almost the entire salient region (see Figure~\ref{fig5} for some examples).

To achieve more imperceptible attacks, we then seek to further restrict the perturbations to even smaller regions.
The intuition is that even in the salient region, some sub-regions are more critical.
For example, it has been shown that the dog face region is the brightest in the salient region of a dog image \cite{21}, and obscuring it will dramatically change the model output \cite{26}.
Furthermore, by combining internal feature visualization with output prediction, it has been revealed in \cite{27} that within dog face region, the ears and eyes seem to be more important than others.
Therefore, it is reasonable to assume the salient region of an image is progressive with respect to its impact on model output.
If we could find smaller but more salient sub-region, the perturbations will be more efficient, leading to more imperceptible attacks (see the third column of Figure~\ref{fig1}).
Thus, we propose the Saliency Attack, a novel black-box attack that recursively refines the perturbations in the salient region.

It is worth mentioning that except white-box attack JSMA, the idea of restricting perturbations to a small region has also been implemented in transfer-based attacks \cite{19,20}, where class activation mapping (CAM) \cite{21} and Grad-CAM \cite{22} are adopted to generate the saliency maps.
However, transfer-based attacks assume the data distribution for training the target model is available and thus could build a substitute model to approximate it, which actually belong to the grey-box setting where partial knowledge of the target model is known.
As a result, they cannot be applied to the more strict black-box setting where only the model output is available (see Section~\ref{sec:2.3} for details).

In summary, we make the following contributions in this work.
\begin{itemize}
	\item To the best of our knowledge, for crafting AEs in the black-box setting, we are the \textit{first} to restrict perturbations to a salient region. This approach is readily compatible with many existing black-box attacks and significantly improves their imperceptibility with little degradation in success rate.
	\item We propose the Saliency Attack, a novel gradient-free black-box attack that iteratively refines perturbations in salient regions to keep them minimal and essential.
	Compared with the SOTA black-box attacks, our approach achieves much better attack imperceptibility in terms of most apparent distortion (MAD), $L_0$ and $L_2$ distances, and also obtains significantly higher success rates judged by a human-like threshold on MAD.
	\item We demonstrate that the perturbations generated by Saliency Attack are more robust against detection-based defenses, including Feature Squeezing and binary classifier detection.
\end{itemize}

The rest of this paper is organized as follows: Section~\ref{sec:2} presents a literature review on black-box attacks. Next, in Section~\ref{sec:3}, we define the optimization problem of our attack and detail the proposed Saliency Attack. Section~\ref{sec:4} shows the experimental results and analysis, followed by a conclusion in Section~\ref{sec:5}.

\section{Related Work}\label{sec:2}
In this part, we first overview the recent work of black-box attacks. Based on the generation method of AEs, these attacks can be divided into gradient estimation attacks and gradient-free attacks. Besides, we also introduce some work related to the imperceptibility in adversarial attacks. Finally, we discuss some methods that extract salient regions in an image.

\subsection{Black-box Attacks}
\subsubsection{Gradient Estimation Attacks}
Gradient estimation attacks first estimate the gradients by querying the target model and then use them to run white-box attacks. ZOO attack \cite{30} first adopts symmetric difference quotient to approximate the gradients and then performs white-box Carlini-Wagner (CW) attack \cite{10}. AutoZOOM \cite{31}, a variant of ZOO, uses a random vector based gradient estimation to reduce the query number per iteration from 2\textit{D} in ZOO to \textit{N}+1 (\textit{D} is the dimensionality and \textit{N} is the sample size). To further enhance query efficiency, Ilyas et al. \cite{14} propose the “tiling trick” that updates a square of pixels simultaneously instead of a single pixel. This dramatically decreases the dimensionality by a factor of $k^2$ ($k$ is tile length).

\subsubsection{Gradient-free Attacks}
Gradient-free attacks do not estimate gradients and directly generate AEs with search heuristics according to the query result. Su et al. \cite{32} propose One-pixel attack that adopts differential evolution algorithm to perturb the most important pixel in the image. Alzantot et al. \cite{33} propose GenAttack, which uses genetic algorithm to generate AEs. To improve query efficiency, Moon et al. \cite{15} consider a discrete surrogate optimization problem that transforms the original constraint of a continuous range $[-\epsilon, +\epsilon]$ to a discrete set $\{-\epsilon, +\epsilon\}$, achieving a massive reduction in the search space. This is motivated by linear program (LP) where the optimal solution is attained at an extreme point of the feasible set \cite{34}. Combing tiling trick \citep{14} and discrete optimization \citep{15}, Square Attack \cite{16} has obtained the best result on success rate and query performance so far with random search.

\subsection{Attack Imperceptibility}
The imperceptibility of AEs is essential for practical attackers, which has been investigated by previous studies from different perspectives.
Guo et al. \cite{43} consider low frequency perturbations as imperceptible, thus searching for AEs in frequency domain.
But Zhang et al. \cite{44} regard imperceptibility as visual smoothness in an image and integrate Laplacian smoothing into optimization.
Croce and Hein \cite{49} use a combination of $L_0$ and $L_{\infty}$ norms to generate sparse and imperceptible perturbations.
Besides, some studies also leverage color distance \cite{50} and image quality assessment (IQA) \cite{51,47} to improve attack imperceptibility.

On the other hand, determining how to assess the imperceptibility of AEs is still an open question.
Most existing adversarial attacks use $L_p$ norms ($L_0$, $L_2$ and $L_\infty$) to measure the human perceptual distance between the perturbed image and the original one.
Nonetheless, it has been shown that $L_p$ norms are not suitable enough for human vision system \cite{46}.
A recent study in \cite{38} systematically examines various IQA metrics including $L_p$ norms through large-scale human evaluation on the imperceptibility of different AEs, and finds that among all the metrics the most apparent distortion (MAD) \cite{39} metric is closest to subjective scores (see Appendix~\ref{appendix:sec:MAD} for details of MAD). Hence, in this research, we adopt MAD as our main metric to assess attack imperceptibility.

\subsection{Extracting Salient Region}\label{sec:2.3}
As aforementioned, there have been white-box attacks or transfer-based attacks that restrict the perturbations to a small salient region.
Specifically, white-box attack JSMA \cite{9} constructs a BP-saliency map by calculating derivatives of the model output w.r.t input pixels \cite{18}, while the two transfer-based attacks \cite{19,20} utilize CAM and Grad-CAM to extract salient regions, respectively. CAM \cite{21} replaces the final fully connected layers with convolutional layers and global average pooling of a CNN, and localizes class-specific salient regions through forward propagation. Grad-CAM \cite{22} improves CAM with on need to modify the network architecture, but it still requires access to the inner parameters of the model to calculate the gradients.
Thus, CAM and Grad-CAM can only be applied to white-box attacks or transfer-based attacks, which first construct a transparent substitute model, craft AEs with gradient-based white-box attacks, and then transfer the generated AEs to the target model. It is conceivable that for transfer-based attacks, the transferability of both AEs and saliency maps highly depends on the similarity between the substitute model and the target model, which further depends on the prior knowledge of the training data distribution. Unfortunately, neither such knowledge nor model gradients are available in black-box settings. Considering the limitations of the above methods, we adopt a salient object detection model \cite{23} to directly generate the saliency map given an input image with no need to access the target model's architecture or parameters.
% Different from prvious methods, salient object detection ...
\section{Proposed Method}\label{sec:3}
\begin{figure}[tbp]
	\centering
	\includegraphics[width=\textwidth]{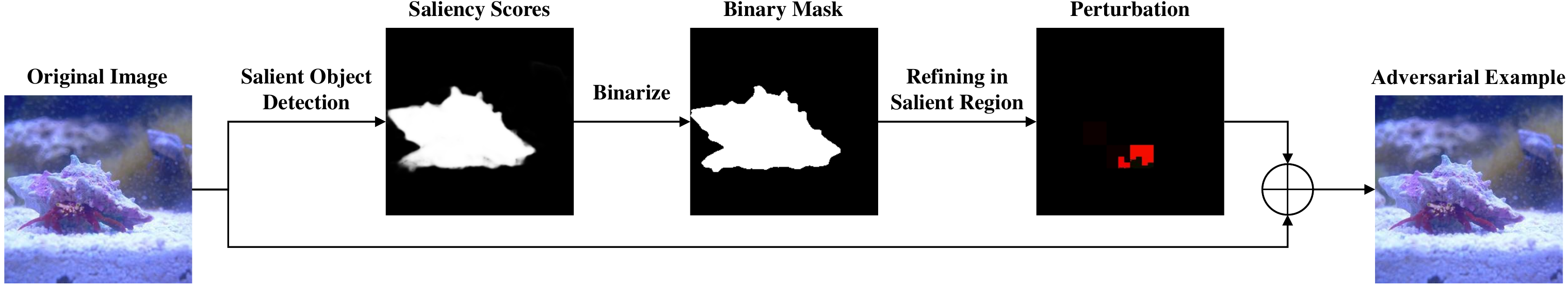}
  % \vspace{-0.7\baselineskip}
	\caption{The overall flowchart of the proposed Saliency Attack.}
	\label{fig3}
\end{figure}

In this section, we first formulate the problem of crafting AEs for image classification models, and then detail our approach.
Figure~\ref{fig3} illustrates the overall flowchart of Saliency Attack.

\subsection{Preliminary}
Given a well-trained DNN classifier $h: [0,1]^{d} \rightarrow \mathbb{R}^{K}$, where $d$ is the dimension of the input $x$, and $K$ is the number of classes.
Let $h_{k}(x)$ denote the predicted score that $x$ belongs to class $k$.
The classifier assigns the class that maximizes $h_{k}(x)$ to the input $x$.
The goal of an untargeted attack is to find an AE $x_{adv}$, that results in the model misclassification from the ground-truth class $y_{gt}$, and meanwhile keeps the distance between $x_{adv}$ and the benign input $x$ smaller than a threshold $\epsilon$:
\begin{equation} \label{eq1}
  \mathop{\arg\max}_{k=1,...,K}\ h_{k}(x_{adv}) \neq y_{gt}, \quad s.t. \ ||x_{adv}-x||_{p} \le \epsilon, \ x_{adv}\in [0,1]^{d}
\end{equation}
Note the last constraint indicates that $x_{adv}$ is a valid image.
In this study, we focus on $L_\infty$ norm, following \cite{15,16}.
Conventionally, the task of finding $x_{adv}$ can be rephrased as solving a constrained continuous problem:
\begin{equation} \label{eq2}
  % \mathop{\max}_{x_{adv} \in [0,1]^d} L(F(x_{adv}), y), \quad s.t. \ ||x_{adv}-x||_{\infty} \le \epsilon
  \mathop{\max}_{x_{adv} \in [0,1]^d}\ f(x_{adv}), \quad s.t. \ ||x_{adv}-x||_{\infty} \le \epsilon
\end{equation}
% where $L$ is a loss function.
where $f(x)=L(x,y_{gt})$ is the loss function.
Similar to \cite{15}, we transform the continuous problem into a discrete surrogate problem, where the perturbation $x_{adv} - x = \delta \in \{-\epsilon, +\epsilon, 0\}^{d}$. Note unlike \cite{15}, where all pixels are perturbed either by $-\epsilon$ or $+\epsilon$, we allow the pixels to remain unperturbed to refrain from global perturbation.
Besides, only the pixels in salient regions can be perturbed. The final problem is defined as the following set maximization problem.
\begin{equation} \label{eq3}
  \mathop{\max}_{\mathcal{S}^{+}, \mathcal{S}^{-} \subseteq \mathcal{S} \atop \mathcal{S}^{+} \cap \mathcal{S}^{-} = \emptyset}\ \left\{F(\mathcal{S}^{+},\mathcal{S}^{-}) \triangleq f\left(x+\epsilon \sum_{i\in \mathcal{S}^{+}}e_i - \epsilon \sum_{i \in \mathcal{S}^{-}}e_i\right)\right\}
\end{equation}
where $\mathcal{V}$ denotes the ground set which is the set of all pixel locations ($|\mathcal{V}|=d$), $\mathcal{S}$ denotes the set of pixels in salient regions, $\mathcal{S}^{+}$ and $\mathcal{S}^{-}$ denote the set of \textit{selected} pixels with $+\epsilon$ and $-\epsilon$ perturbations respectively, and $e_i$ is the $i$-th standard basis vector. Note that $\mathcal{S} \subseteq \mathcal{V}$, and $\mathcal{S}\setminus (\mathcal{S}^{+} \cup \mathcal{S}^{-})$ is the set of pixels in salient regions that are unperturbed.
The goal of Equation (\ref{eq3}) is to find $\mathcal{S}^{+}$ and $\mathcal{S}^{-}$ that will maximize the objective set function $F$.

% \begin{equation} \label{eq3}
%   \mathop{\max}_{x_{adv} \in [0,1]^d} L(F(x_{adv}), y), \quad s.t. \ \delta \in \{-\epsilon, +\epsilon, 0\}^{\mathcal{S}}
% \end{equation}
% where $\mathcal{S}$ is the set of the pixels in salient region.

\subsection{Salient Object Detection}
Salient object detection aims to automatically and accurately extract salient object(s) in an image.
Compared with other salient region extraction methods like BP-saliency map \cite{18}, CAM \cite{21} and Grad-CAM \cite{22}, this type of models do not require any information other than the input image, which is very suitable for the black-box setting.
We adopt the Pyramid Feature Attention (PFA) network\footnote{Implementation of PFA: https://github.com/sairajk/PyTorch-Pyramid-Feature-Attention-Network-for-Saliency-Detection} \cite{23} that achieves the SOTA performance on multiple datasets via capturing high-level context features and low-level spatial structural features simultaneously (see Appendix~\ref{appendix:sec:PFA} for details). Specifically, given an input image, PFA can generate a saliency score between 0 and 1 for each pixel. The higher value denotes higher visual saliency. We then use a threshold $\phi$ to transform the saliency scores into a binary saliency mask, which determines the salient region.
The binarization can be expressed as
\begin{equation} \label{eq4}
  s^{*}_{i}=\begin{cases}
  0 \qquad s_{i} < \phi \\
  1 \qquad s_{i} \ge \phi \\
\end{cases}
\end{equation}
where $s_{i}$ and $s^{*}_{i}$ are the saliency score and the binary mask at the $i$-th pixel position, respectively. Thus, the salient region is the set of the pixels masked by 1.
\begin{equation} \label{eq5}
  \mathcal{S} \triangleq \left\{i\ |\ s^{*}_{i} = 1,\ i \in \mathcal{V} \right\}
\end{equation}

\subsection{Refining Perturbations in Salient Region}

\begin{figure}[h]
  \centering
  \includegraphics[width=0.45\textwidth]{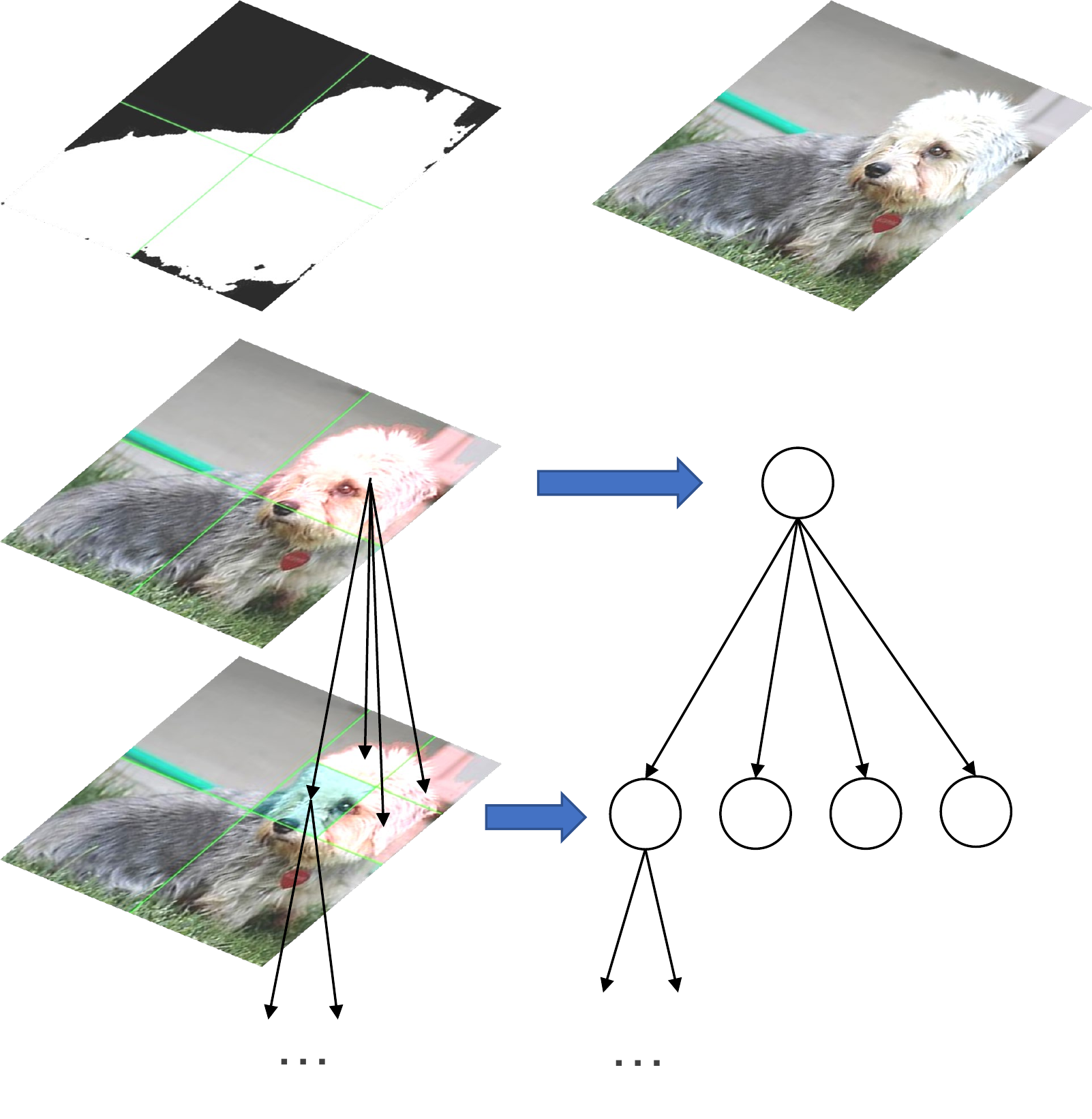}
  % \vspace{-1\baselineskip}
  \caption{The illustration of refining on a tree structure of the image blocks. Among initial blocks, each block is a root node of a tree, and its child nodes are its finer blocks. For each block, only the parts in salient regions will be perturbed.
}\label{fig4}
\end{figure}

The proposed Saliency Attack is outlined in Algorithm~\ref{alg1}. The search process (lines 6-35) recursively refines perturbations based on a tree structure for an image, which is illustrated in Figure~\ref{fig4}.
Concretely, an input image is first split into some initial blocks (a coarse grid in Figure~\ref{fig4}), and only the blocks in salient regions are kept (lines 7-11).
Then we try to add $+\epsilon$ or $-\epsilon$ perturbations on each initial block individually (lines 12-19) and sort them according to their influence ($F$) on model output (line 26).
After that, we choose the block with most influence for further refinement if its $F$ is better than the current best $\hat{F}$ (lines 27-35).
We then recursively refine the current best block (line 32), split this block into smaller blocks (a finer grid in Figure~\ref{fig4}), and again try to add a perturbation on each block individually (at this time we just flip the perturbation (lines 21-24) for convenience, e.g., $+\epsilon$ to $-\epsilon$) to find the best smaller block.
The refinement process recurs until the minimal block (e.g., 1 pixel, line 30) or no smaller block has a better $F$ (line 28). Afterwards, we backtrack to the last level of split blocks and use the second-best block for further perturbation.
In this way, the most important block will be explored first, and the perturbations could be as small as possible, with no need to initialize a global perturbation like Parsimonious Attack~\cite{15}.

To make full use of query budget and combine the advantages of different initial block sizes $k_{int}$ (large $k_{int}$ can quickly lead to successful attacks with large perturbations, while small $k_{int}$ enables the refinement for small perturbations in finer grids), we leverage an outer iteration to run the aforementioned \textit{Refine} search with decreasing $k_{int}$ (lines 2-5). During the iterations, the algorithm will stop if the generated $x_{adv}$ succeeds to fool the model ($F>0$) or the termination condition is reached (line 2).

We use the loss function from CW attack \cite{10} of untargeted attack for Equation (\ref{eq2}):
\begin{equation} \label{eq6}
  L(x, y_{gt})= - \mathop{\max}(Z(x_{adv})_{y_{gt}} - \mathop{\max}_{i \ne y_{gt}}(Z(x_{adv})_{i}), \; 0)
  \vspace{-4pt}
\end{equation}
where $Z(x_{adv})_{y_{gt}}$ is the logit with respect to the ground-truth class of the original image. In this way, the loss function is imposed to leave a margin between the ground-truth class and other classes.
%
% \begin{algorithm}[h]
% 	\caption{Saliency Attack} \label{alg1}
% 	\KwIn{original image $x$, initial block size $k_{int}$, salient region $\mathcal{S}$, query budget}
% 	\KwOut{AE $x_{adv}$}
%   $\delta \gets \emptyset$ is the set of perturbed pixels\;
%   \While{$k_{int}>1$ and not exceeding query budget}{
%     \textit{Refine($x,k_{int}$)}\;
%     $k_{int} \gets k_{int}/2$\;
%     }
%   $x_{adv} \gets$ apply the perturbations $\delta$ to $x$\;
%   \vspace{5pt}
%   \textbf{Procedure} \textit{Refine($b, k$)} \\
%   $k^{\prime}$ is the block size of $b$\;
%   $\{b_{1}, b_{2},...,b_{(k^{\prime}/k)^2}\} \gets$ split $b$ into $(k^{\prime}/k)^2$ finer blocks\;
%   $\{b_{1}, b_{2},...,b_{n}\} \gets \{b_{1}, b_{2},...,b_{(k^{\prime}/k)^2}\} \cap \mathcal{S}$\;
%   $\{b_{\pi(1)}, b_{\pi(2)},...,b_{\pi(n)}\} \gets$ sort $\{b_{1}, b_{2},...,b_{n}\}$ in descending order according to $L(b)$, which is calculated by using Equation~(\ref{eq6})\;
%   \For{each block $e \in \{b_{\pi(1)}, b_{\pi(2)},...,b_{\pi(n)}\}$}{
%     \If{$L(e) > \hat{L}$}{
%       $\delta \gets \delta \cup \{e\}$\;/-*
%       $\hat{L} \gets L(e)$\;
%       \If{$k > 1$}{
%         $k \gets k/2$\;
%         Recursively call \textit{Refine($e,k$)}\;
%         }
%       }
%     }
% \end{algorithm}

\begin{algorithm}[h]
  \small
	\caption{Saliency Attack} \label{alg1}
	\KwIn{Objective set function $F$, Ground set $\mathcal{V}$, salient region $\mathcal{S}$,  initial block size $k_{int}$, query budget}
	\KwOut{$\mathcal{S}^{+}, \mathcal{S}^{-}$ set of pixels with $+\epsilon$ and $-\epsilon$ perturbations}
  $\mathcal{S}^{+} \gets \emptyset$; $\mathcal{S}^{-} \gets \emptyset$; $\hat{F} \gets -\infty$\;
  \While{$k_{int}>1$ and not exceeding query budget}{
    \textit{Refine($\mathcal{V},\ k_{int},\ 1$)}\;
    $k_{int} \gets k_{int}/2$\;
  }
  % \vspace{5pt}
  \textbf{Procedure} \textit{Refine($B,\ k,\ split\_level$)} \\
  \tcc{Split the current block into finer blocks in the salient region.}
  $k^{\prime} \gets \sqrt{|B|}$ is the block size of $B$; $n \gets (k^{\prime}/k)^2$ is the number of split blocks\;
  $\{B_{1}, B_{2},...,B_{n}\} \gets$ split $B$ into $n$ finer blocks\;
  \For{$i\leftarrow 1$ \KwTo $n$}{
    $B_{i} \gets B_{i} \cap \mathcal{S}$\;
  }
  \tcc{Sort the finer blocks according to their influence on model output.}
  \uIf{$split\_level=1$}{
    \For{$j\leftarrow 1$ \KwTo $n$}{
      \uIf{$F(\mathcal{S}^{+}\cup B_{j}, \mathcal{S}^{-}\setminus B_{j}) \ge F(\mathcal{S}^{+}\setminus B_{j}, \mathcal{S}^{-}\cup B_{j})$}{
        $\mathcal{S}^{+}_{j} \gets \mathcal{S}^{+}\cup B_{j}$;\ $\mathcal{S}^{-}_{j} \gets \mathcal{S}^{-}\setminus B_{j}$;\ $F_{j} \gets F(\mathcal{S}^{+}\cup B_{j}, \mathcal{S}^{-}\setminus B_{j})$\;
      }
      \Else{
        $\mathcal{S}^{+}_{j} \gets \mathcal{S}^{+}\setminus B_{j}$;\ $\mathcal{S}^{-}_{j} \gets \mathcal{S}^{-}\cup B_{j}$;\ $F_{j} \gets F(\mathcal{S}^{+}\setminus B_{j}, \mathcal{S}^{-}\cup B_{j})$\;
      }
    }
  }
  \Else{
    \For{$j\leftarrow 1$ \KwTo $n$}{
      $\mathcal{S}^{+}_{j} \gets (\mathcal{S}^{+}\setminus B_{j})\cup(\mathcal{S}^{-}\cap B_{j})$;\ $\mathcal{S}^{-}_{j} \gets (\mathcal{S}^{-}\setminus B_{j})\cup(\mathcal{S}^{+}\cap B_{j})$\;
      $F_{j} \gets F(\mathcal{S}^{+}, \mathcal{S}^{-})$\;
    }
  }
  $\{F_{\pi(1)}, F_{\pi(2)},..., F_{\pi(n)}\} \gets$ sort $\{F_{1}, F_{2},..., F_{n}\}$ in descending order\;
  \tcc{Recursively select the finer block with most influence for perturbation.}
  \For{$j\leftarrow 1$ \KwTo $n$}{
    \If{$F_{\pi(j)} > \hat{F}$}{
      $\mathcal{S}^{+}\gets \mathcal{S}^{+}_{\pi(j)}$;\ $\mathcal{S}^{-}\gets \mathcal{S}^{-}_{\pi(j)}$;\ $\hat{F}\gets F_{\pi(j)}$\;
      \If{$k > 1$}{
        $k \gets k/2$\;
        Recursively call \textit{Refine($B_{\pi(j)},\ k,\ split\_level+1$)}\;
      }
    }
  }
\end{algorithm}

\section{Experiments}\label{sec:4}
The main goal of the experiments is to validate: \textit{(1) whether salient regions could improve the imperceptibility of existing black-box attacks; (2) whether our Saliency Attack could further enhance the imperceptibility performance.}
Hence, we first compare our Saliency Attack with the baselines, including SOTA black-box attacks and their modified version restricted in salient regions. Then we perform ablation studies to verify the effectiveness of salient regions and \textit{Refine} search separately. Besides, we measure the hyperparameter sensitivity of our approach. Finally, we test different detection-based defenses to evaluate the imperceptibility from the defensive perspective.

\subsection{Settings}
We compare our Saliency Attack against SOTA black-box attacks including Boundary Attack \cite{45}, TVDBA \cite{47}, Parsimonious attack \cite{15} and Square attack \cite{16}. Among them, Boundary Attack can constantly reduce the perturbation to a very small magnitude with enough queries, TVDBA tries to minimize the perturbation via integrating Structural SIMilarity (SSIM) \cite{48} into the loss, Parsimonious Attack proposes discrete optimization to improve query efficiency, and Square Attack can achieve the current SOTA query efficiency and success rate in the black-box attack setting.
In addition, we design Parsimonious-sal Attack and Square-sal Attack as baselines that restrict their perturbations in salient regions.
For our Saliency Attack, the threshold $\phi$ to produce binary saliency masks is chosen to be 0.1. For splitting, original images are resized to $256 \times 256$, and we set the initial block size $k_{int}$ to 16. All parameters of the compared attacks remain consistent with those recommended in their papers.

For performance metrics, we employ commonly used success rate (SR) and average number of queries (Avg. queries). To evaluate the imperceptibility of AEs, we consider $L_0$, $L_2$ and MAD. All these three imperceptibility metrics are the smaller the better. In practice, a successful AE with imperceptible perturbation is what we need. So besides $\rm SR$, we use a new metric $\rm SR_{true}$. It indicates the rate of successful AEs whose MAD values are below a human-like threshold, i.e., 30, that AEs with $\rm MAD \le 30$ are basically imperceptible to human eyes (See Appendix~\ref{appendix:sec:threshold}).

In this study, we consider $L_{\infty}$ threat model on ImageNet dataset \cite{1}, and set the perturbation magnitude $\epsilon$ to 0.05 in [0,1] scale.
We use Inception v3 \cite{41} as the target model, and test different query budgets \{3,000, 10,000, 30,000\} for untargeted attack. We provide our implementation publicly\footnote{https://github.com/Daizy97/SaliencyAttack}.

\begin{figure}[ht]
  \centering
  \scalebox{1.0}{
  \includegraphics[width=\textwidth]{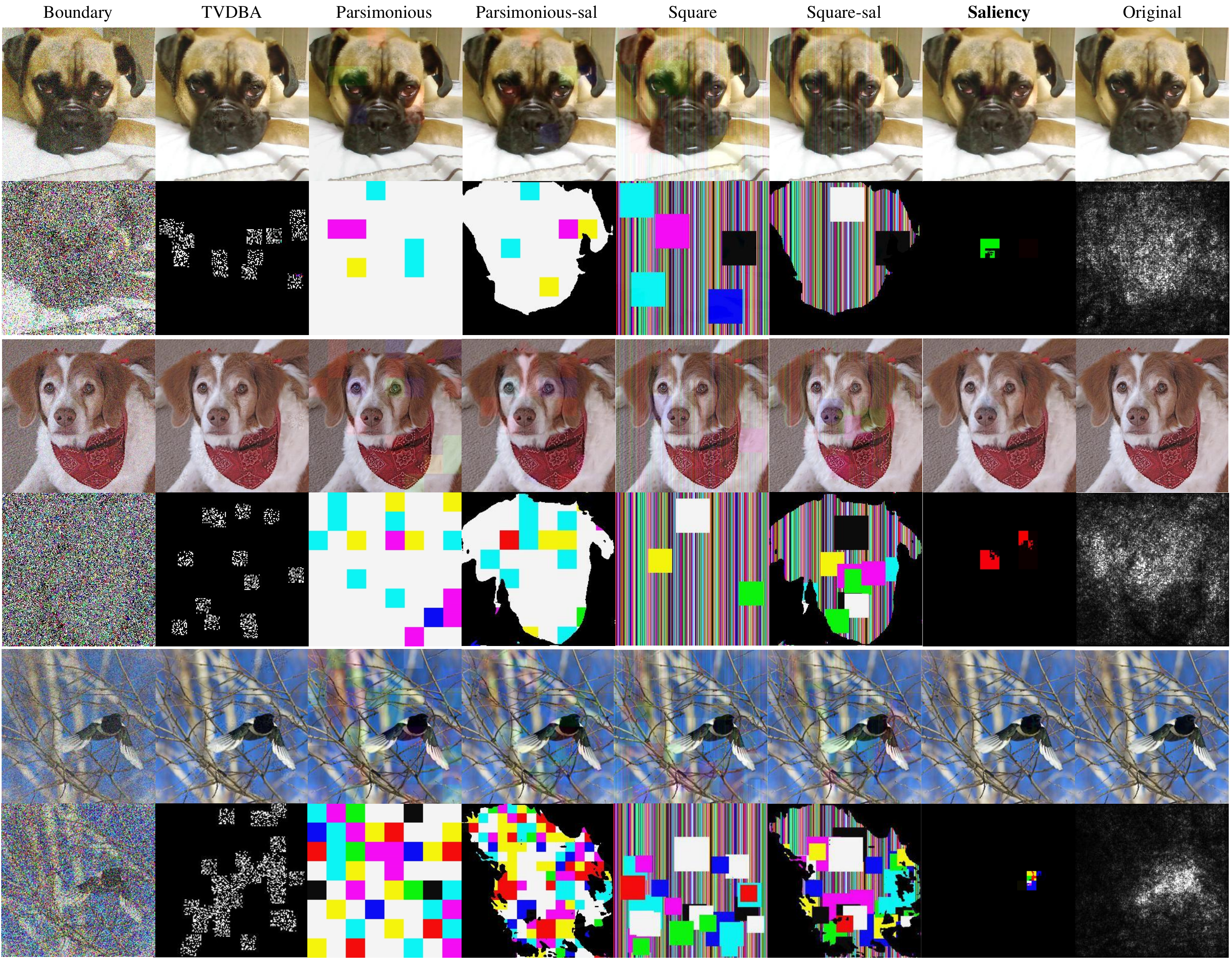}}
  % \vspace{-1.5\baselineskip}
  \caption{Examples generated by different attacks. For each example, the upper row is AEs and original image. The lower row is perturbations and BP-saliency map.}
  \label{fig5}
\end{figure}

\subsection{Results and Analysis}
% Our salient region can be easily applied to existing attacks, so we also design Parsimonious-sal Attack and Square-sal Attack as baselines that restrict their perturbations in salient regions.
Some examples are shown in Figure~\ref{fig5}. We can easily find the perturbations of original Parsimonious Attack and Square Attack are very obvious in the entire image due to global perturbation, while their modified versions are relatively more imperceptible since no perturbation exists in background regions. However, their perturbations are still complex and irregular, taking up almost all salient regions. In comparison, even restricted in the same salient regions, the perturbations generated by Saliency Attack are smaller and more critical, roughly corresponding to the bright pixels in BP-saliency map. They further represent the positions of dogs' noses or ears, which accord with our inspiration before. Besides, the AEs of Boundary Attack contain noticeable coarse textures due to inadequate query budget, and TVDBA produces global and irregular perturbations similarly.

\begin{table}[t]
	\caption{Comparison of different attacks. $\rm SR_{true}$ denotes the rate of successful AEs with $\rm MAD \le 30$. $\rm SD$ denotes the standard deviation calculated over 10 groups of total 10,000 AEs. The best results are in bold based on the Wilcoxon signed-rank test with significance level at 0.05.} \label{tab2}
	\centering
  \scalebox{0.9}{
	\begin{tabular}{c|cccccc}
		\toprule
		{Method} & {Budget} & {$\rm SR\pm SD$} & {$\rm SR_{true}\pm SD$} & {$\rm L_2\pm SD$} & {$\rm L_0\pm SD$} & {$\rm MAD\pm SD$}\\
		\midrule
    \multirow[c]{3}{*}{Boundary} %& 1000 & 100.0\% & 6.5\% & 79.82 & 99.4\% & 97.56\\
    & 3,000 & 100.0\%$\pm$0.00 & 13.6\%$\pm$0.04 & 68.13$\pm$0.74 & 99.6\%$\pm$0.00 & 84.96$\pm$4.19\\
    & 10,000 & 100.0\%$\pm$0.00 & 30.2\%$\pm$0.03 & 58.82$\pm$0.65 & 99.7\%$\pm$0.00 & 71.61$\pm$4.03\\
    & 30,000 & 100.0\%$\pm$0.00 & 46.0\%$\pm$0.03 & 58.30$\pm$0.62 & 99.8\%$\pm$0.00 & 62.77$\pm$3.94\\
    \midrule
    \multirow[c]{3}{*}{TVDBA} %& 1000 & 100.0\% & 6.5\% & 79.82 & 99.4\% & 97.56\\
    & 3,000 & 90.1\%$\pm$0.01 & 23.4\%$\pm$0.02 & 10.64$\pm$0.15 & 19.0\%$\pm$0.00 & 37.65$\pm$3.63\\
    & 10,000 & 96.9\%$\pm$0.01 & 23.7\%$\pm$0.02 & 10.83$\pm$0.18 & 19.8\%$\pm$0.00 & 37.87$\pm$3.61\\
    & 30,000 & 98.4\%$\pm$0.00 & 23.5\%$\pm$0.02 & 10.93$\pm$0.17 & 20.0\%$\pm$0.00 & 38.11$\pm$3.57\\
    \midrule
    \multirow[c]{3}{*}{Parsimonious} %& 1,000 & 80.7\% & 14.0\% & 22.17 & 100.0\% & 45.59\\
    & 3,000 & 92.4\%$\pm$0.01 & 14.7\%$\pm$0.01 & 22.17$\pm$0.00 & 100.0\%$\pm$0.00 & 51.80$\pm$0.12\\
    & 10,000 & 98.5\%$\pm$0.01 & 14.8\%$\pm$0.01 & 22.17$\pm$0.00 & 100.0\%$\pm$0.00 & 53.01$\pm$0.14\\
    & 30,000 & 99.8\%$\pm$0.01 & 14.8\%$\pm$0.01 & 22.17$\pm$0.00 & 100.0\%$\pm$0.00 & 53.13$\pm$0.14\\
    \midrule
    \multirow[c]{3}{*}{Parsimonious-sal} %& 1,000 & 76.2\% & 9.9\% & 14.11 & 42.7\% & 45.18\\
    & 3,000 & 88.9\%$\pm$0.01 & 11.4\%$\pm$0.01 & 13.54$\pm$0.50 & 35.5\%$\pm$0.03 & 45.31$\pm$0.21\\
    & 10,000 & 96.0\%$\pm$0.01 & 12.6\%$\pm$0.01 & 13.47$\pm$0.48 & 35.5\%$\pm$0.03 & 45.44$\pm$0.22\\
    & 30,000 & 99.2\%$\pm$0.01 & 12.9\%$\pm$0.01 & 13.45$\pm$0.48 & 35.5\%$\pm$0.03 & 45.28$\pm$0.21\\
    \midrule
    \multirow[c]{3}{*}{Square} %& 1000 & 91.5\% & 2.4\% & 25.37 & 99.1\% & 58.00\\
    & 3000 & 98.3\%$\pm$0.00 & 2.7\%$\pm$0.00 & 25.34$\pm$0.02 & 99.0\%$\pm$0.00 & 57.29$\pm$0.05\\
    & 10,000 & 99.6\%$\pm$0.00 & 2.4\%$\pm$0.00 & 25.35$\pm$0.02 & 99.0\%$\pm$0.00 & 57.74$\pm$0.05\\
    & 30,000 & 99.8\%$\pm$0.00 & 3.1\%$\pm$0.00 & 25.33$\pm$0.02 & 99.0\%$\pm$0.00 & 56.21$\pm$0.04\\
    \midrule
    \multirow[c]{3}{*}{Square-sal} %& 1000 & 73.2\% & 18.1\% & 16.66 & 42.5\% & 39.63\\
    & 3,000 & 87.5\%$\pm$0.01 & 20.3\%$\pm$0.02 & 14.71$\pm$0.53 & 34.5\%$\pm$0.03 & 38.56$\pm$0.54\\
    & 10,000 & 96.0\%$\pm$0.01 & 22.8\%$\pm$0.02 & 14.56$\pm$0.51 & 34.3\%$\pm$0.03 & 38.65$\pm$0.56\\
    & 30,000 & 96.4\%$\pm$0.01 & 25.9\%$\pm$0.01 & 14.48$\pm$0.51 & 34.1\% $\pm$0.03& 37.23$\pm$0.53\\
    \midrule
    \multirow[c]{3}{*}{Saliency (ours)} %& 1000 & 58.4\% & \textbf{54.7\%} & \textbf{3.57} & \textbf{3.0\%} & \textbf{11.42}\\
    & 3,000 & 84.3\%$\pm$0.01 & \textbf{79.8\%$\pm$0.01} & \textbf{3.44$\pm$0.05} & \textbf{2.9\%$\pm$0.00} & \textbf{12.07$\pm$0.20}\\
    & 10,000 & 92.4\%$\pm$0.01 & \textbf{85.8\%$\pm$0.01} & \textbf{3.63$\pm$0.05} & \textbf{3.4\%$\pm$0.00} & \textbf{12.62$\pm$0.21} \\
    & 30,000 & 94.8\%$\pm$0.01 & \textbf{87.5\%$\pm$0.01} & \textbf{3.81$\pm$0.06} & \textbf{4.0\%$\pm$0.00} & \textbf{13.01$\pm$0.20} \\
		\bottomrule
	\end{tabular}}
  % \vspace{-14pt}
\end{table}

The quantitative results are reported in Table~\ref{tab2}.
In this experiment, we randomly choose 10,000 images from the ImageNet validation set and divide them into ten groups. Then we calculate the mean and standard deviation (SD). The best results are recorded in bold based on Wilcoxon signed-rank test with significance level at 0.05.
We can find in all query budgets, our Saliency Attack statistically significantly outperforms all the baselines with a huge gap in terms of $\rm SR_{true}$ and three imperceptibility metrics, which demonstrates the superiority of our method to refine the perturbations in salient regions. For baselines, although TVDBA, Parsimonious Attack and Square Attack can obtain a better $\rm SR$, their successful AEs are not truly imperceptible and will be easily detected by some defenses or humans. For Boundary Attack, it can gradually make progress on $\rm SR_{true}$, $L_2$ and MAD as the query budget increases, but they are still much worse than Saliency Attack. This is because it takes at least hundreds of thousands of queries for Boundary Attack to converge \cite{45}, which is infeasible in practice. It is also obvious that Parsimonious-sal Attack and Square-sal Attack obtain better imperceptibility performance than their original version with little degradation in SR. This suggests the use of salient regions for black-box attacks is practicable.

\begin{figure}[h]
 \centering
 \includegraphics[width=0.325\textwidth]{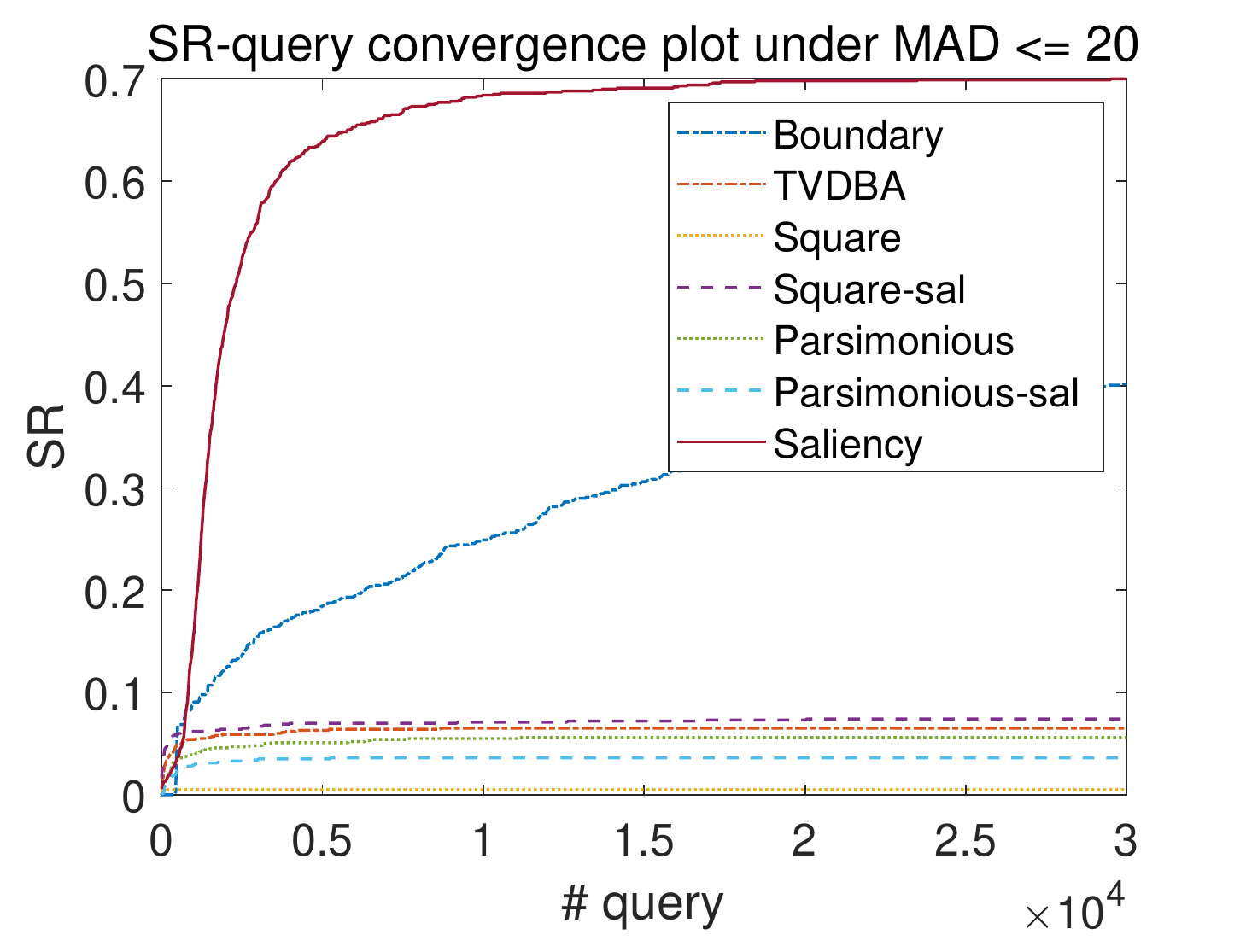}
 \hfill
 \includegraphics[width=0.325\textwidth]{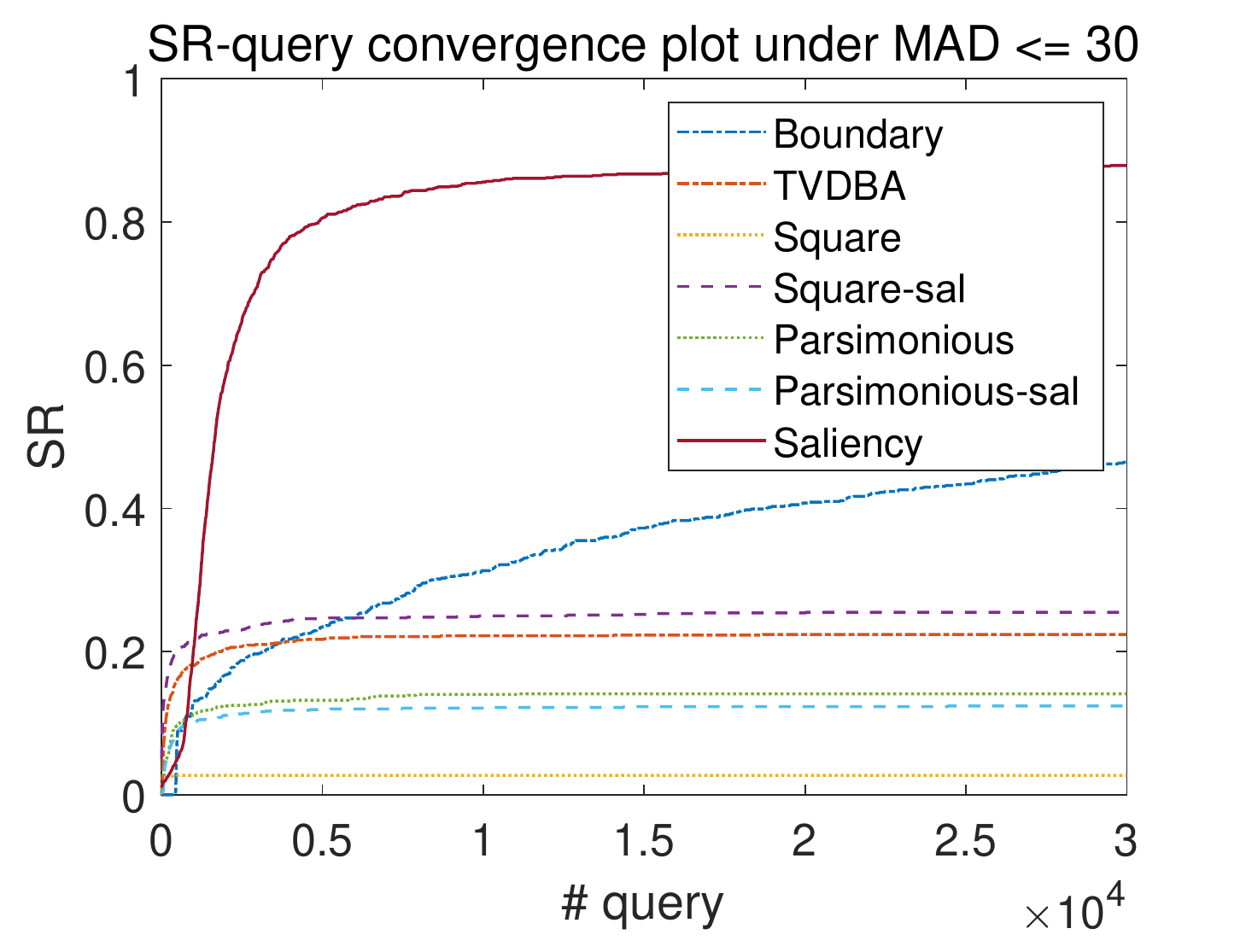}
 \hfill
 \includegraphics[width=0.325\textwidth]{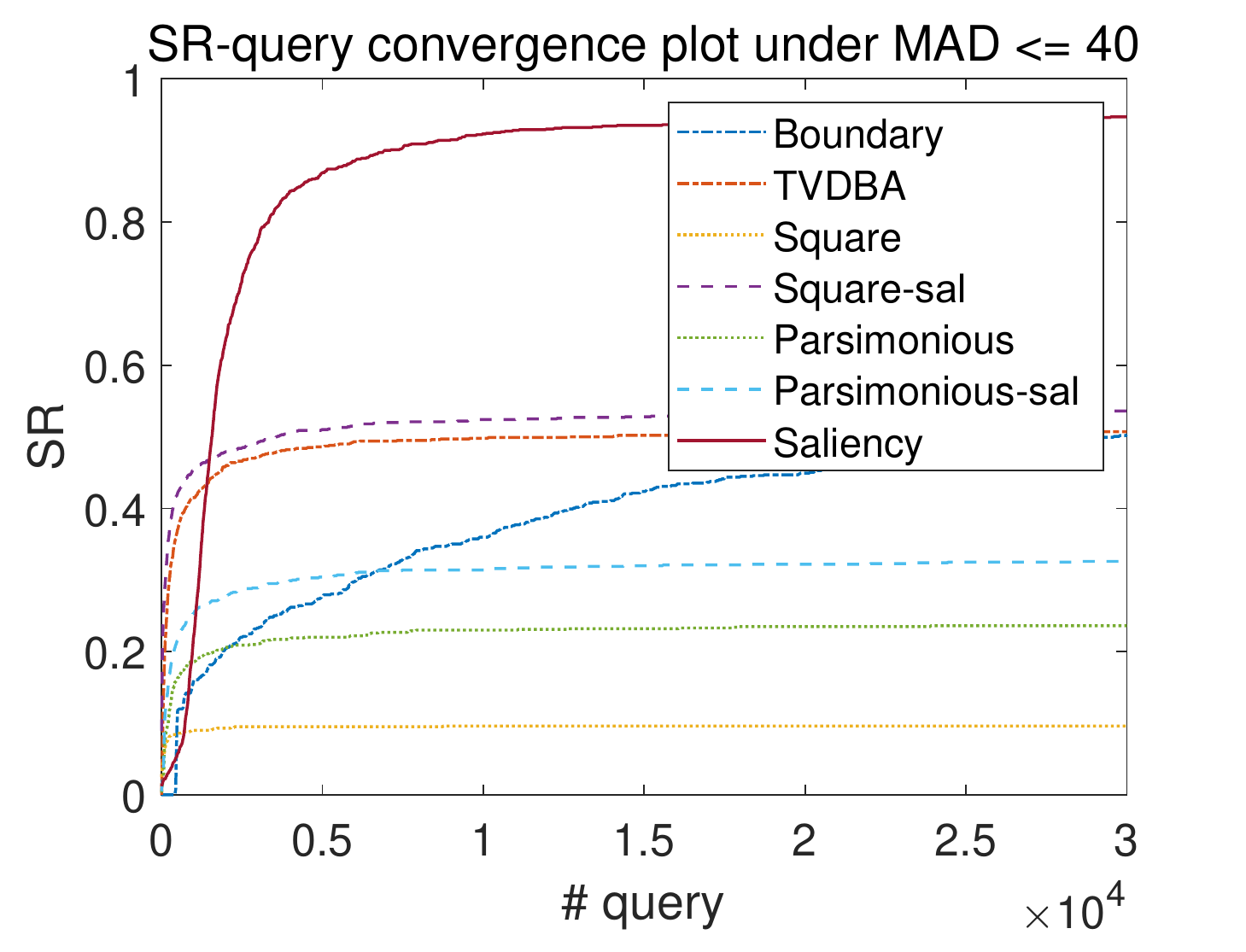}
 \hfill
 % \vspace{-1\baselineskip}
 \caption{Queries vs. True Success Rate under different thresholds of MAD scores}\label{fig6}
\end{figure}

We also plot the convergence curve of $\rm SR_{true}$ versus query number under different MAD thresholds in Figure~\ref{fig6}. It shows that Saliency Attack can lead other attacks stably most of the time. For some baselines such as Square Attack and Parsimonious Attack, they can generate some successful AEs with very few queries due to global perturbations and some tricks like initializing perturbations with random vertical stripes \cite{16}, but they lack the ability to further improve the imperceptibility. Instead, Saliency Attack conservatively selects small regions to perturb, hence its query efficiency is a little lower than some baselines at the beginning but soon exceeds them.
We further study the change of their MAD scores in Appendix~\ref{appendix:sec:scatter} and find most AEs can be improved after restricting in salient region, which also validates the effectiveness of our method.

\subsection{Ablation Study}
\begin{figure}[h]
 \centering
 \subfigure[Salient region]{
   \begin{minipage}[b]{0.14\textwidth}
     \includegraphics[width=\textwidth]{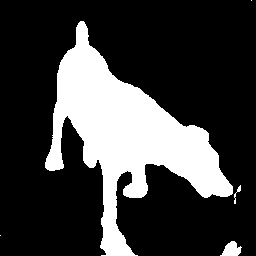}
     \includegraphics[width=\textwidth]{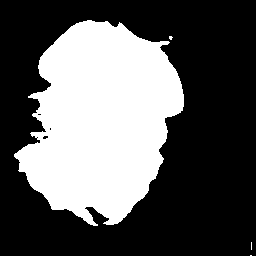}
   \end{minipage}}%
   \hspace{-1.3mm}
 \subfigure[In salient region]{
   \begin{minipage}[b]{0.28\textwidth}
     \includegraphics[width=\textwidth]{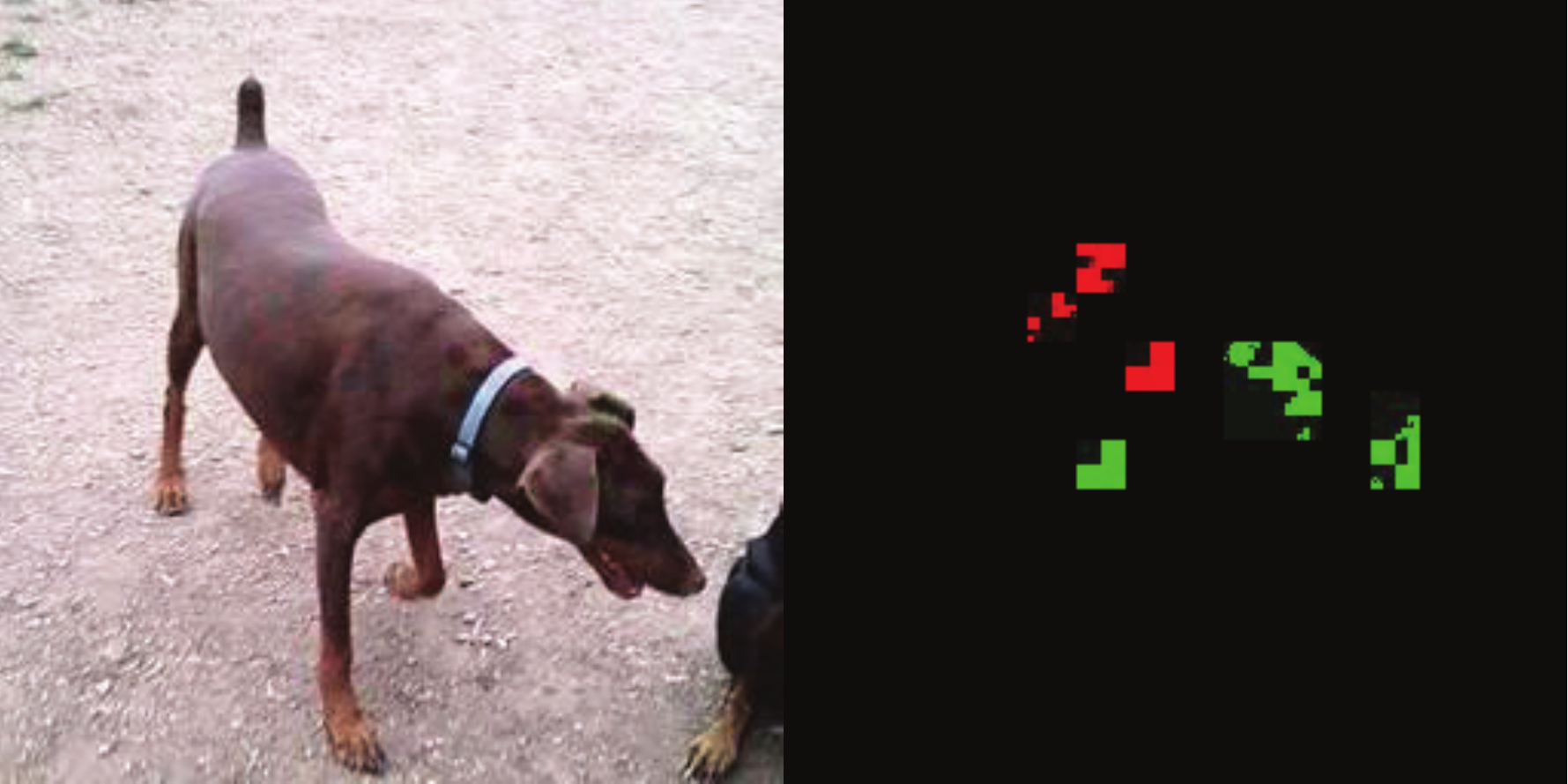}
     \includegraphics[width=\textwidth]{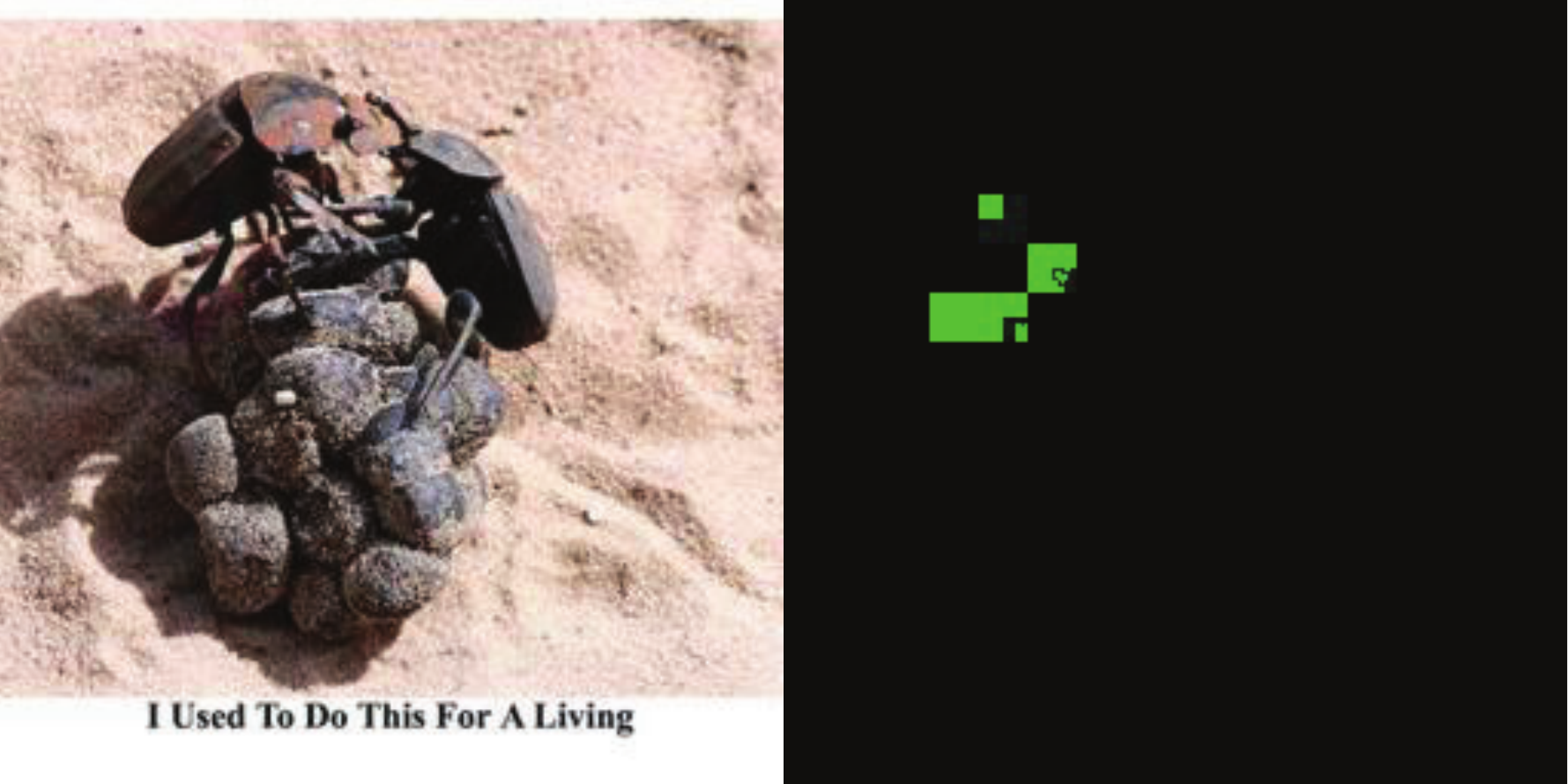}
   \end{minipage}}%
   \hspace{-1.3mm}
 \subfigure[Without saliency]{
   \begin{minipage}[b]{0.28\textwidth}
     \includegraphics[width=\textwidth]{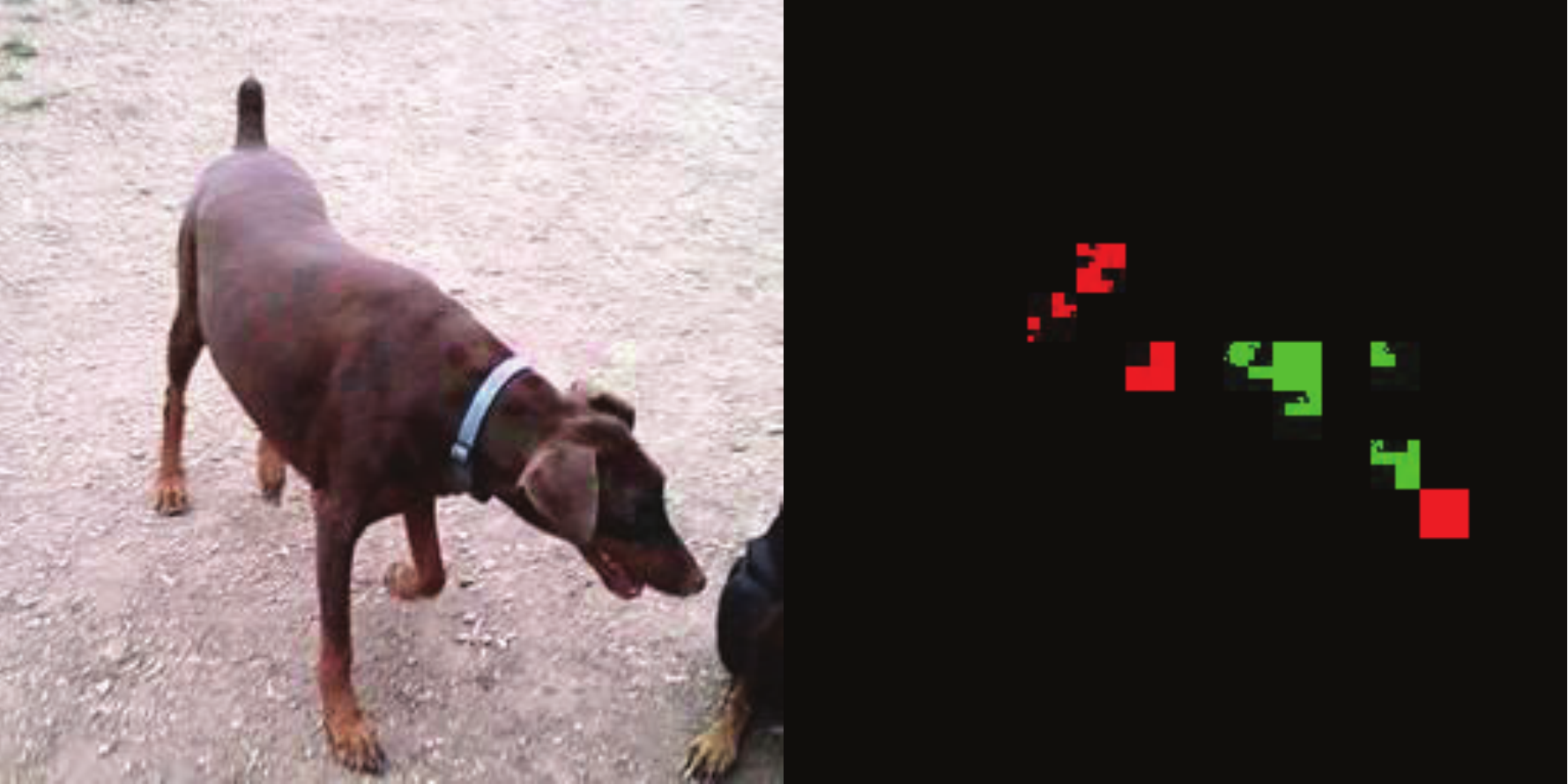}
     \includegraphics[width=\textwidth]{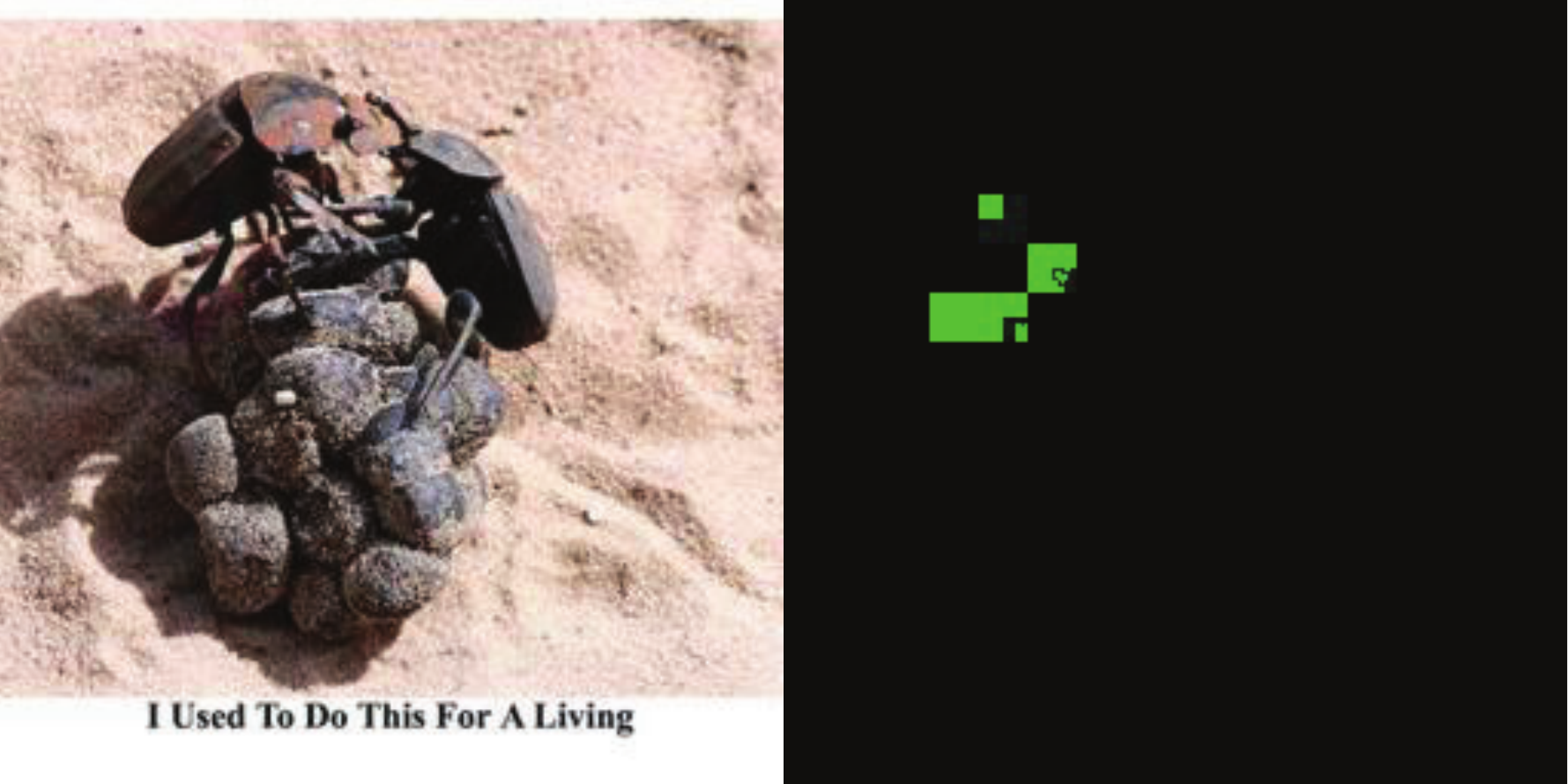}
   \end{minipage}}%
   \hspace{-1.3mm}
 \subfigure[In non-salient region]{
   \begin{minipage}[b]{0.28\textwidth}
     \includegraphics[width=\textwidth]{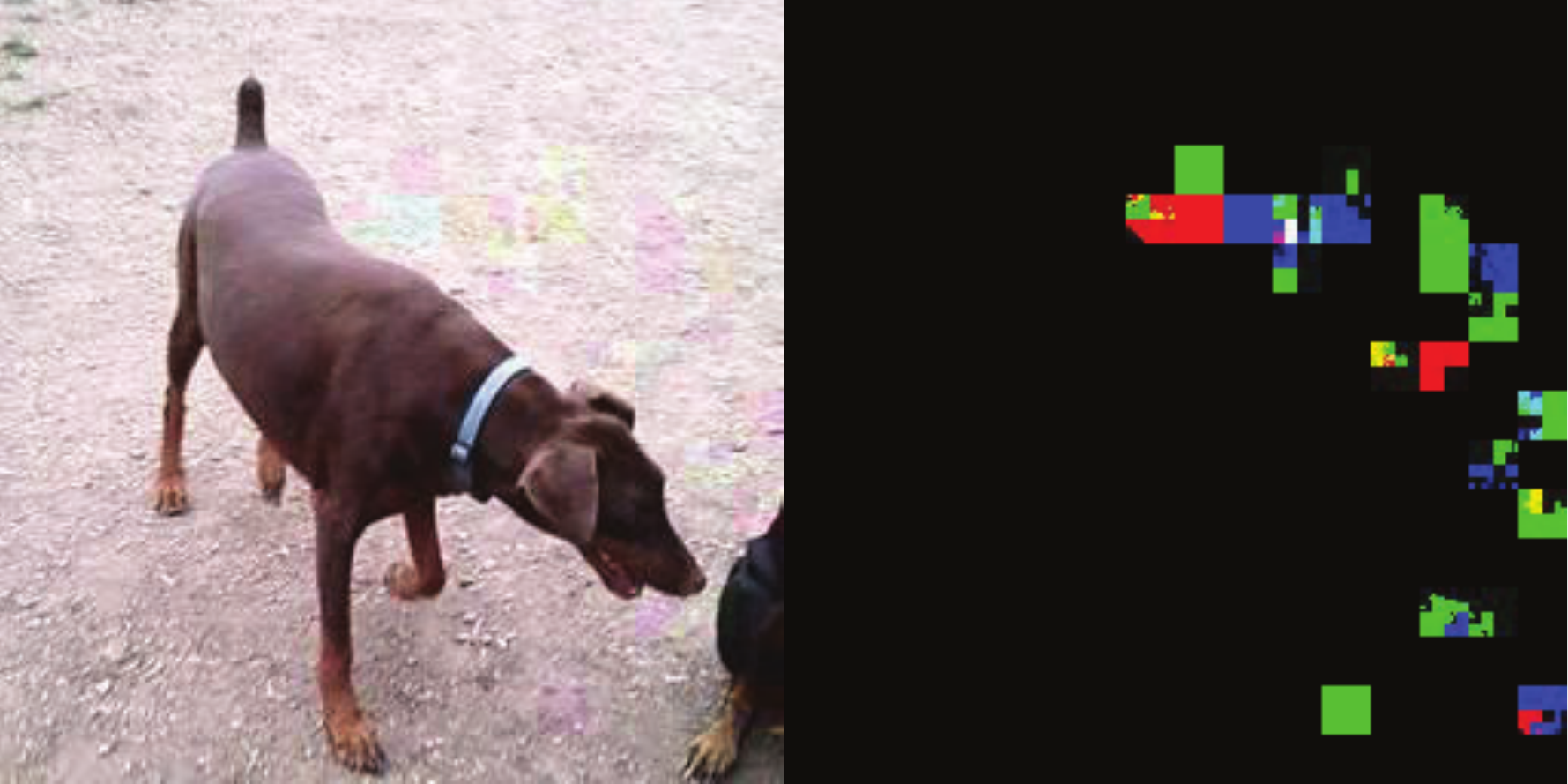}
     \includegraphics[width=\textwidth]{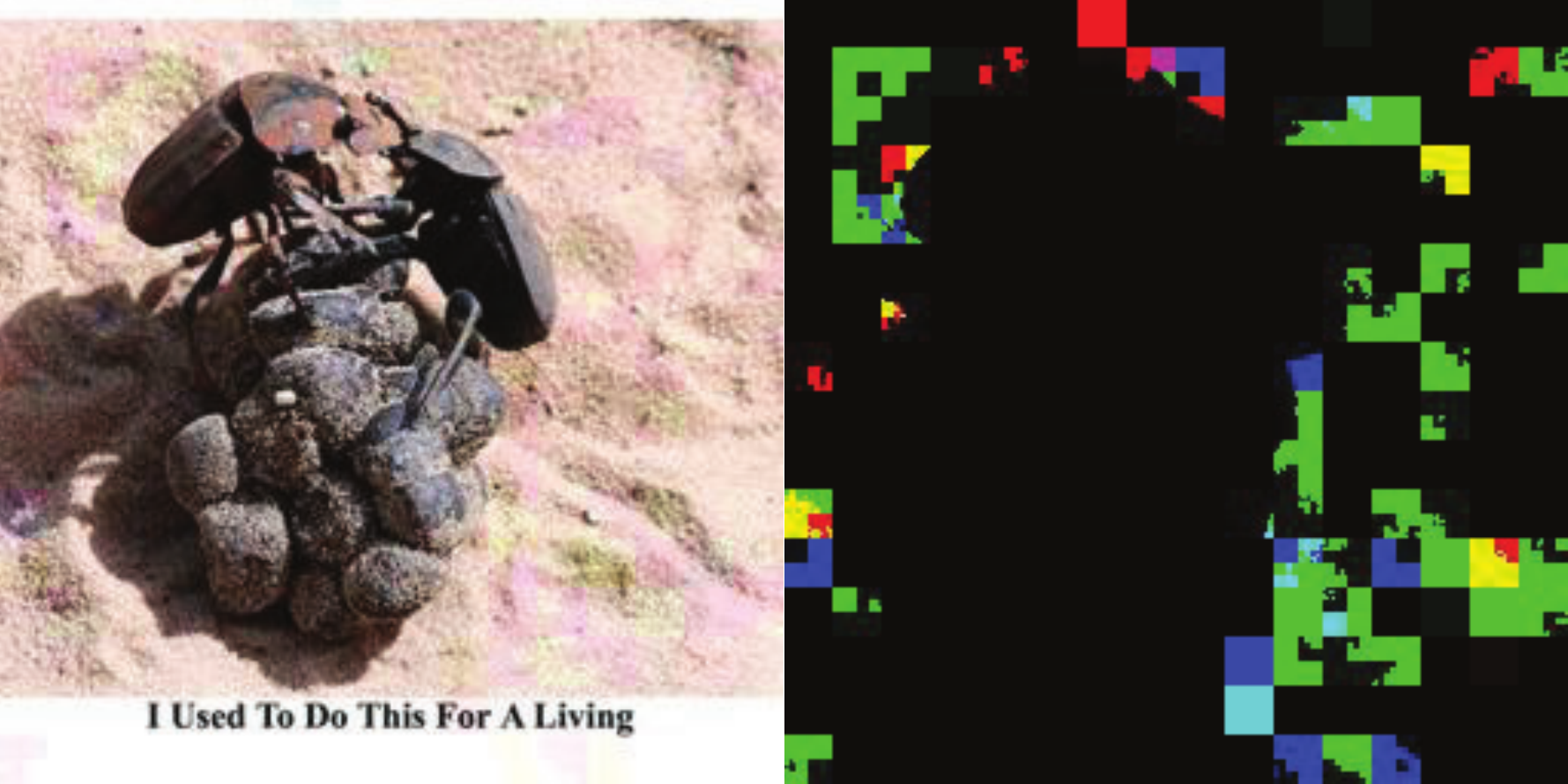}
   \end{minipage}}%
   % \vspace{-1\baselineskip}
 \caption{Examples generated by Saliency Attack with different strategies.}\label{fig10}
\end{figure}

We carry out ablation studies of Saliency Attack, including refining in salient region, in non-salient region, and without saliency (refining in the whole image). We also design a greedy search as a baseline to verify our \textit{Refine} search. We test multiple block sizes for greedy search and use 32 as the best choice (see Appendix~\ref{appendix:sec:GS}). The results on 1,000 randomly selected images and some examples are given in Table~\ref{tab3} and Figure~\ref{fig10}. Note that refining in salient region and refining without saliency generate the same or almost the same perturbations, which means the salient regions indeed contain useful parts and enhance the query efficiency by limiting the search space. But for refining in non-salient region, its perturbation is more complex and visible with worse query efficiency and SR due to unuseful search space. Compared with greedy search, our \textit{Refine} search has much better query efficiency and $\rm SR_{true}$, which demonstrates its superiority. Therefore, we can conclude that both salient region and \textit{Refine} search facilitate Saliency Attack.

\begin{table}[h]
 % \small
	\caption{Ablation study of Saliency attack.} \label{tab3}
	% \vspace{-8pt}
	\centering
	\begin{tabular}{c|cccccc}
		\toprule
   {Method} & {Avg. queries} & {$\rm SR$} & {$\rm SR_{true}$} & {$\rm L_2$} & {$\rm L_0$} & {$\rm MAD$}\\
		\midrule
   {Refining in salient region} & \textbf{1958} & 93.6\% & \textbf{86.2\%} & \textbf{3.71} & \textbf{3.8\%} & \textbf{12.88}\\
   {Refining in non-salient region} & 3128 & 78.2\% & 57.0\% & 4.94 & 6.5\% & 21.58\\
   {Refining without saliency} & 2563 & 95.5\% & 79.6\% & 3.84 & 4.2\% & 16.35\\
   \midrule
   {Greedy search in salient region} & 2727 & 56.0\% & 50.7\% & 4.37 & 4.7\% & \textbf{12.87} \\
		\bottomrule
	\end{tabular}
 % \vspace{-4pt}
\end{table}

\subsection{Hyperparameter Sensitivity}
\begin{figure}[h]
\centering
\includegraphics[width=0.47\textwidth]{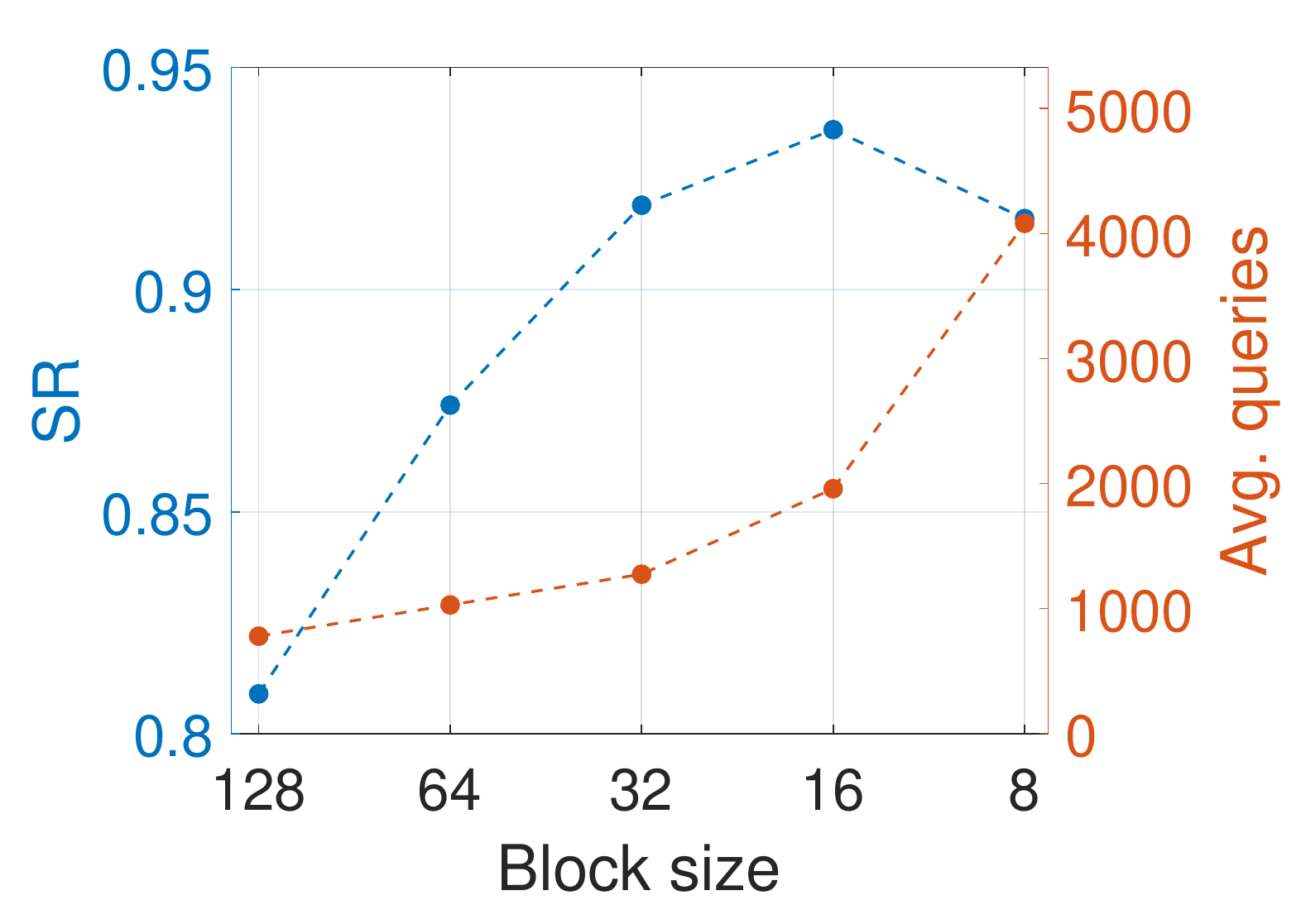}
\hfill
\includegraphics[width=0.47\textwidth]{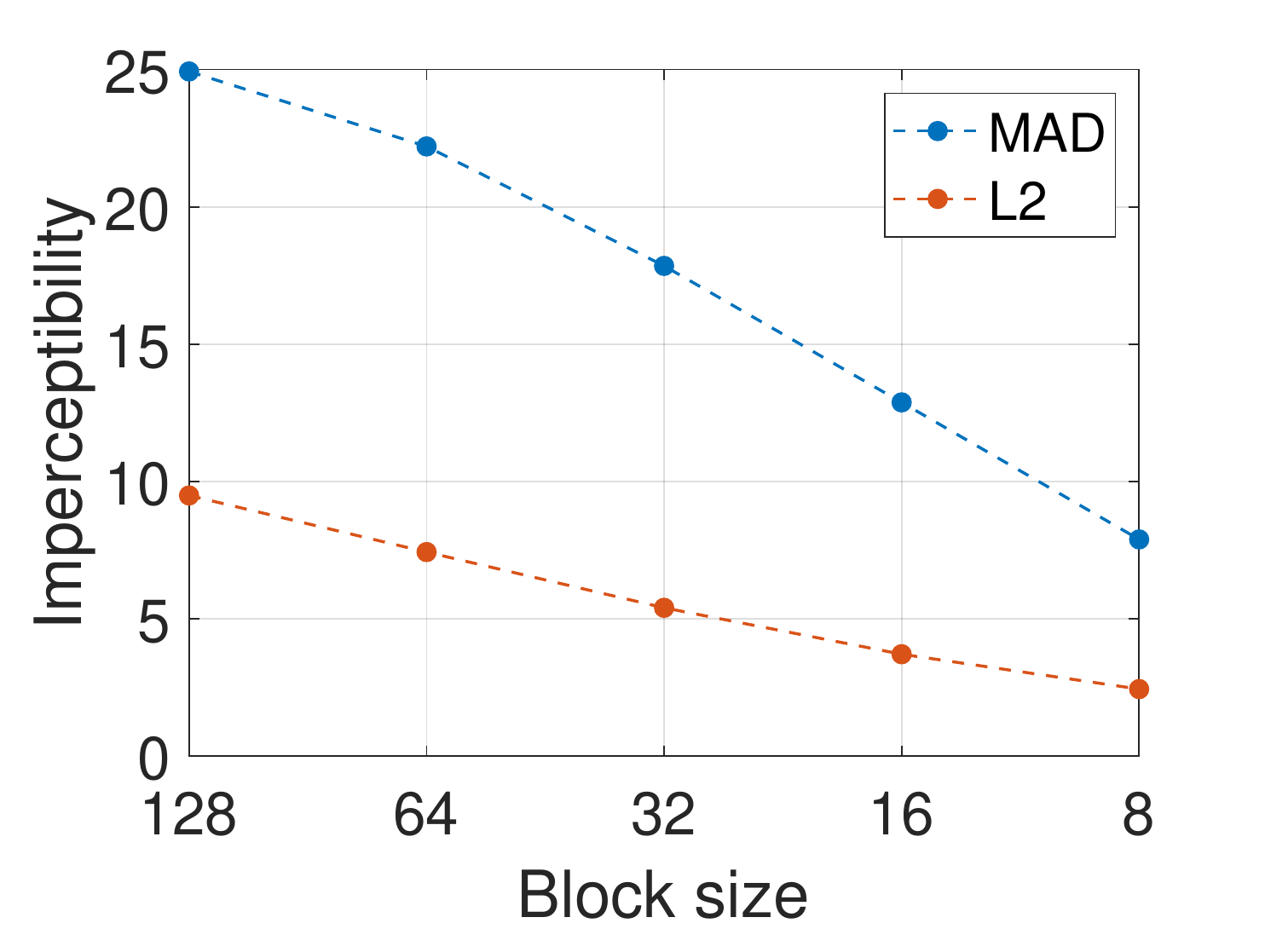}
\hfill
% \vspace{-0.6\baselineskip}
\caption{The effect of initial block size on SR, avg.queries and imperceptibility ($L_2$ and MAD).}\label{fig8}
\end{figure}

The hyperparameter $k_{int}$ is the initial block size that determines the first level of split blocks. We test the effect of different $k_{int}$ on SR, query number, and imperceptibility ($L_2$ and MAD) over 1,000 randomly selected examples in Figure~\ref{fig8}. As $k_{int}$ decreases, SR and imperceptibility can be improved, and SR reaches the peak when $k_{int}$ equals 16. This is because, with smaller initial blocks, Saliency Attack can search for perturbations more finely and accurately leading to higher SR and better imperceptibility. Meanwhile, inevitably more queries are needed, especially for sorting initial blocks. Under a limited query budget like 10,000, the remaining query budget for searching in finer blocks could be inadequate. That's why a turning point occurs in SR.

\begin{figure}[h]
\centering
\subfigure[Salient region]{
  \begin{minipage}[b]{0.12\textwidth}
    \includegraphics[width=\textwidth]{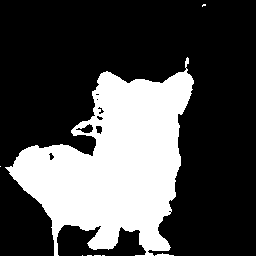}
    \includegraphics[width=\textwidth]{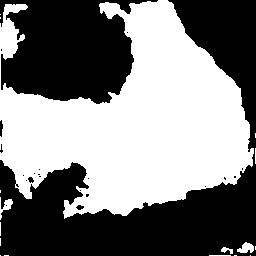}
  \end{minipage}}%
  \hspace{-1.3mm}
\subfigure[$k_{int} = 128$]{
  \begin{minipage}[b]{0.24\textwidth}
    \includegraphics[width=\textwidth]{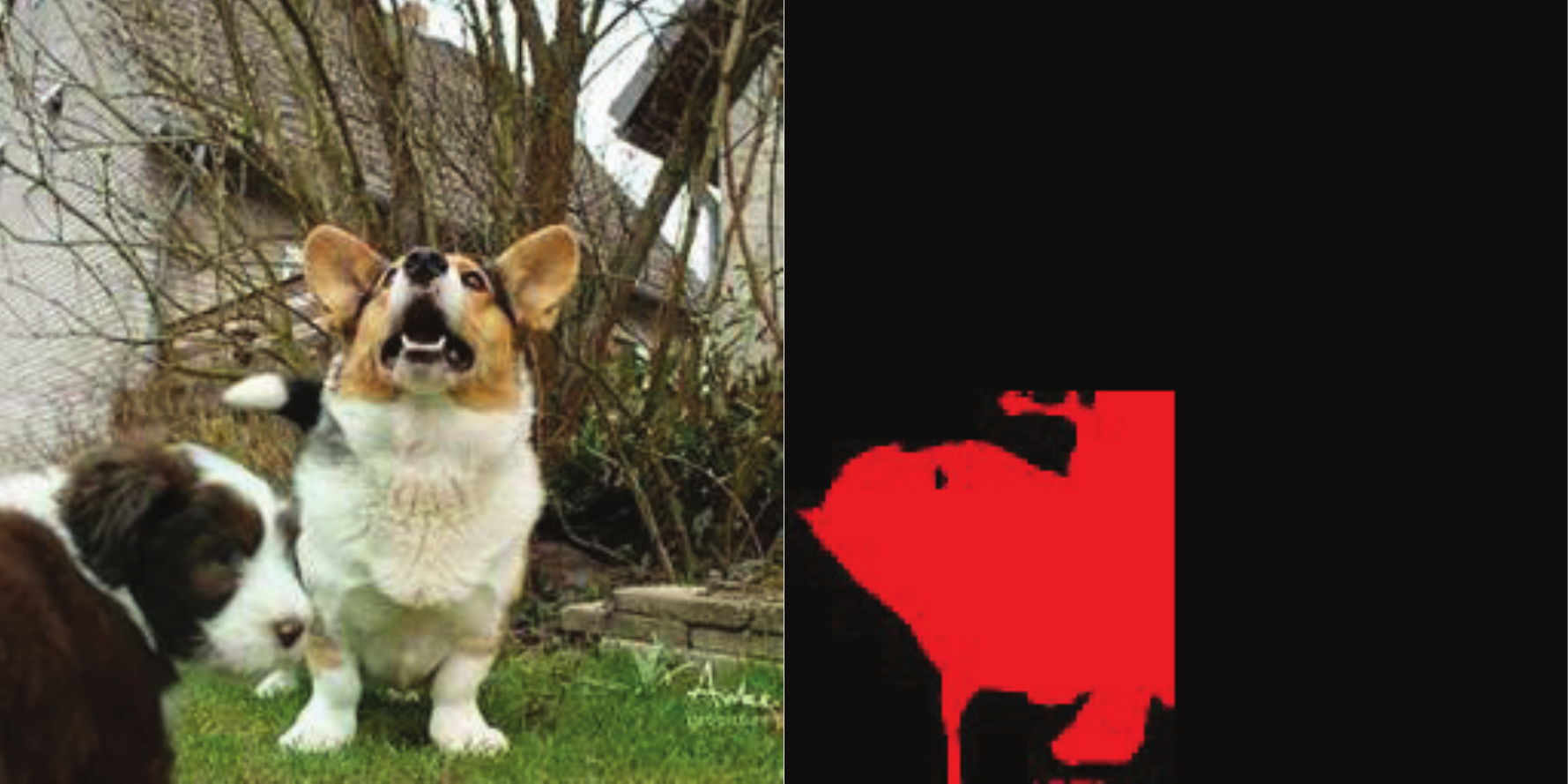}
    \includegraphics[width=\textwidth]{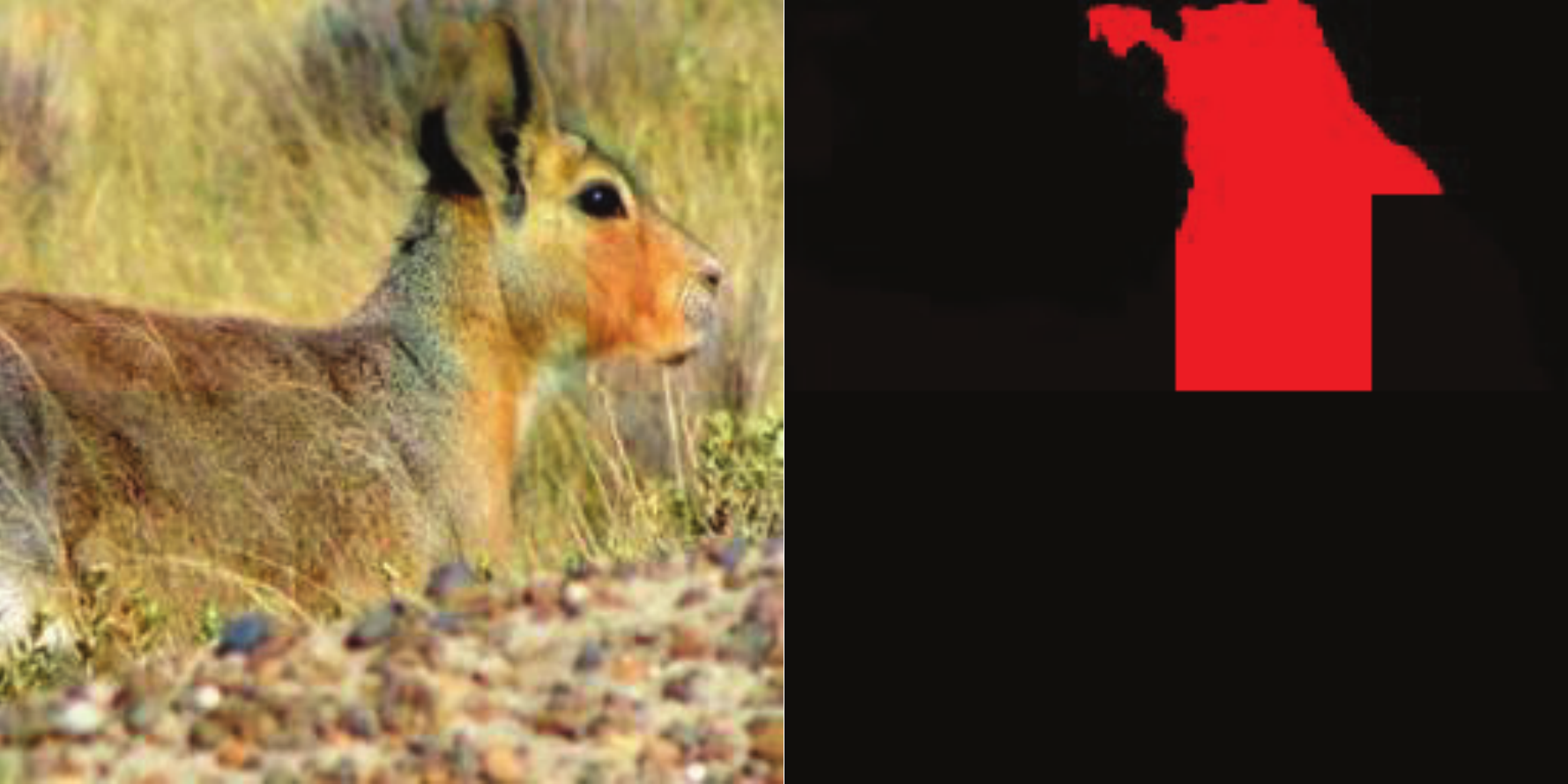}
  \end{minipage}}%
  \hspace{-1.3mm}
\subfigure[$k_{int} = 32$]{
  \begin{minipage}[b]{0.24\textwidth}
    \includegraphics[width=\textwidth]{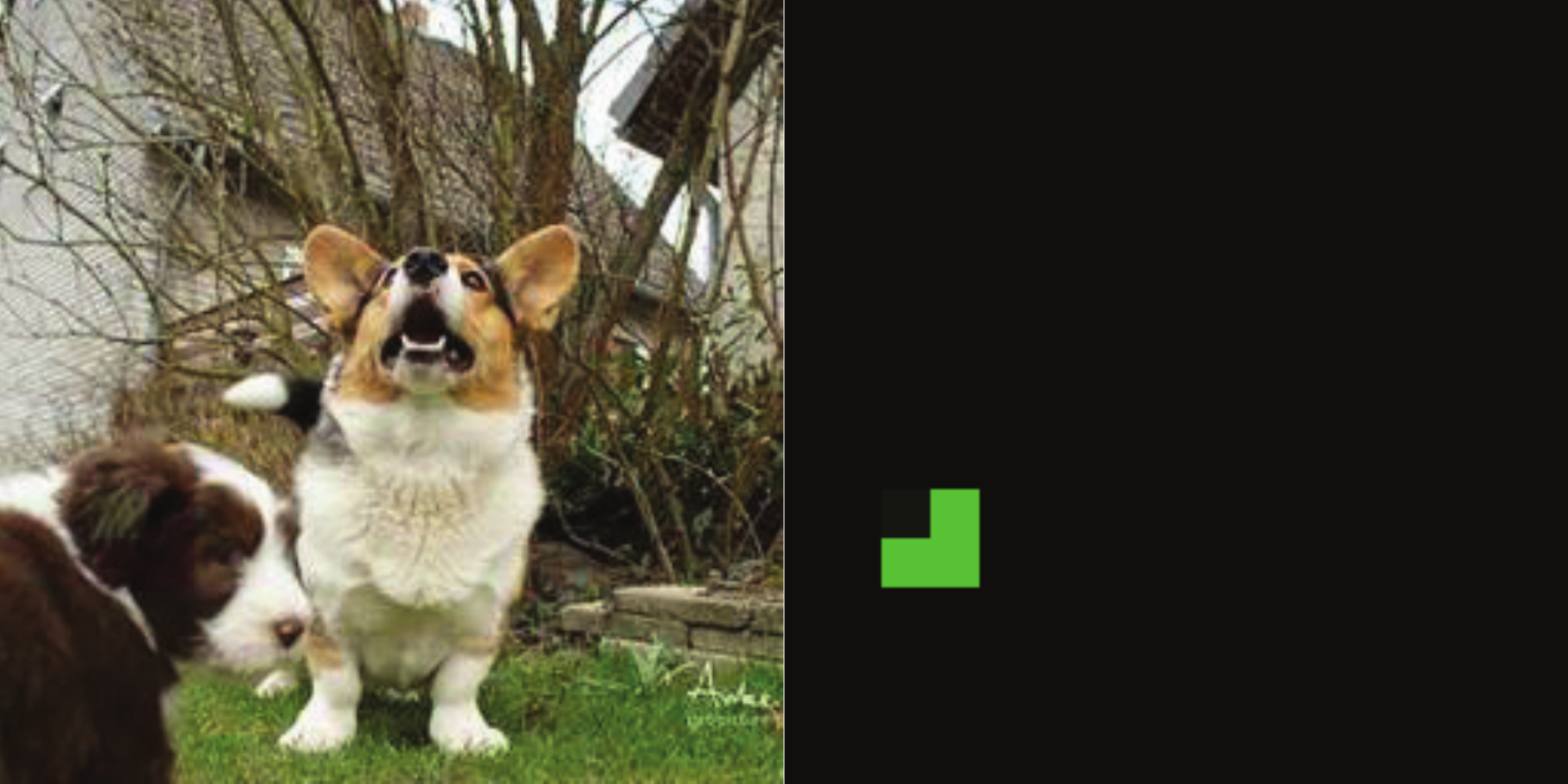}
    \includegraphics[width=\textwidth]{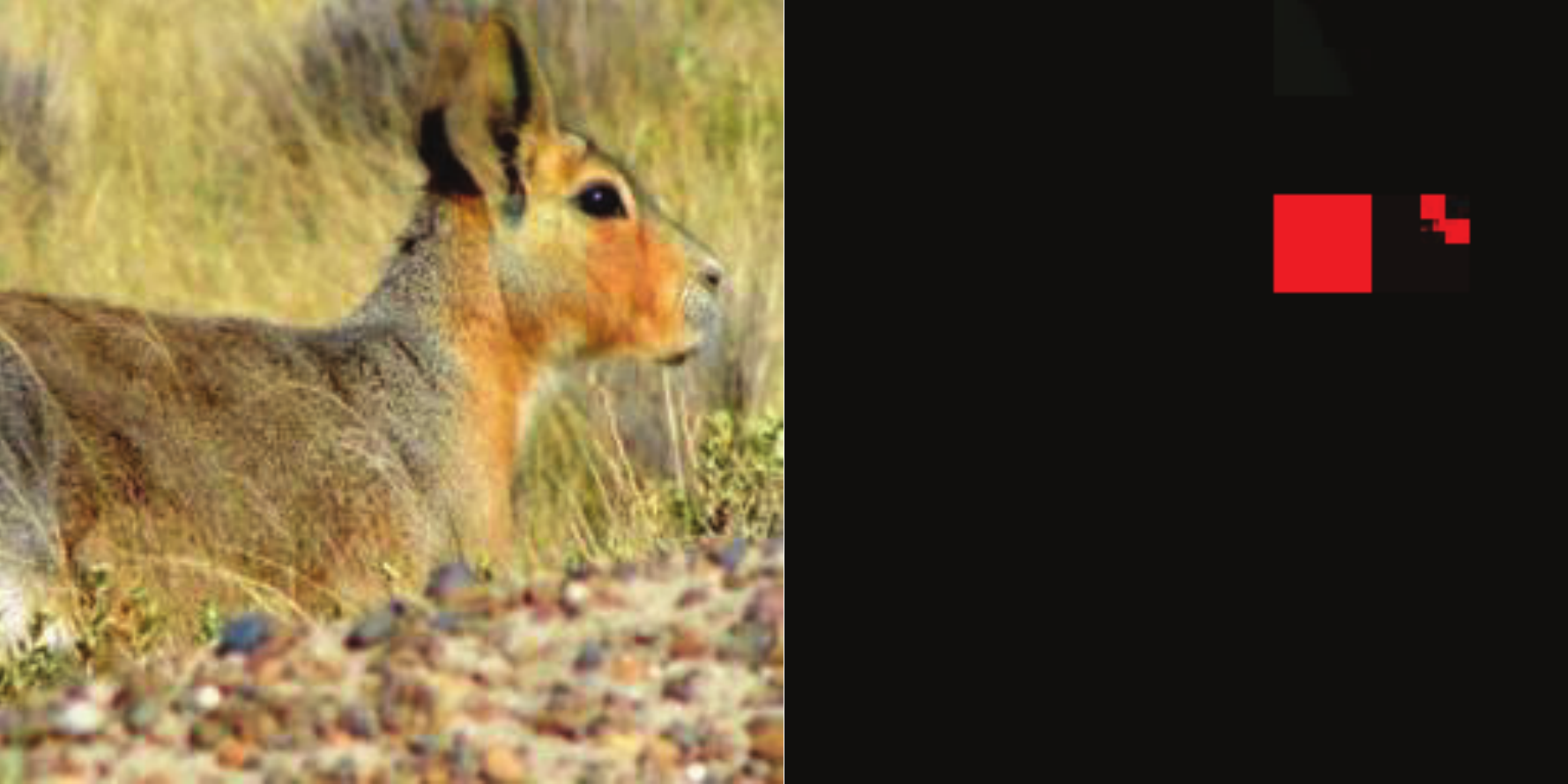}
  \end{minipage}}%
  \hspace{-1.3mm}
\subfigure[$k_{int} = 16$]{
  \begin{minipage}[b]{0.24\textwidth}
    \includegraphics[width=\textwidth]{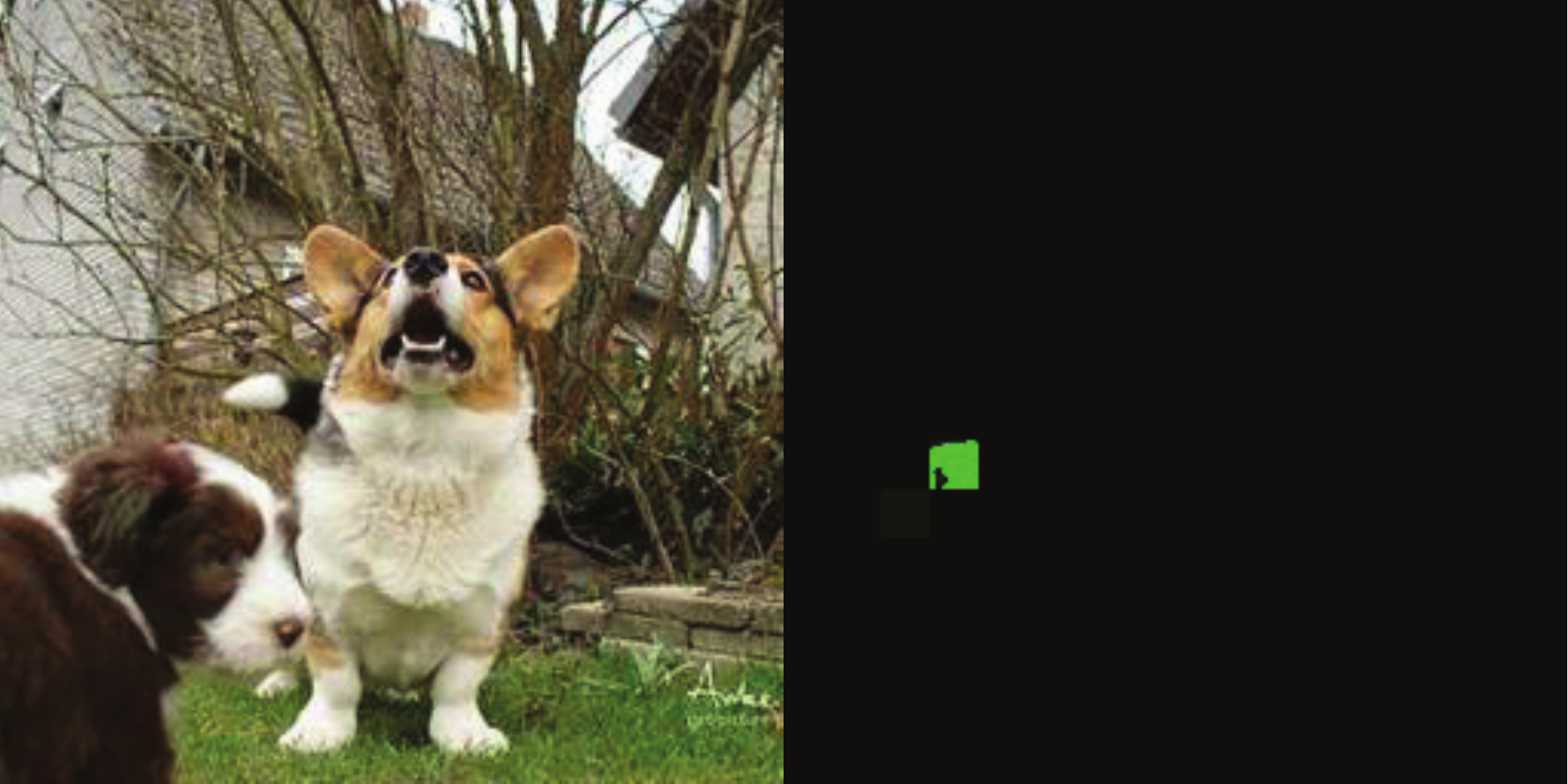}
    \includegraphics[width=\textwidth]{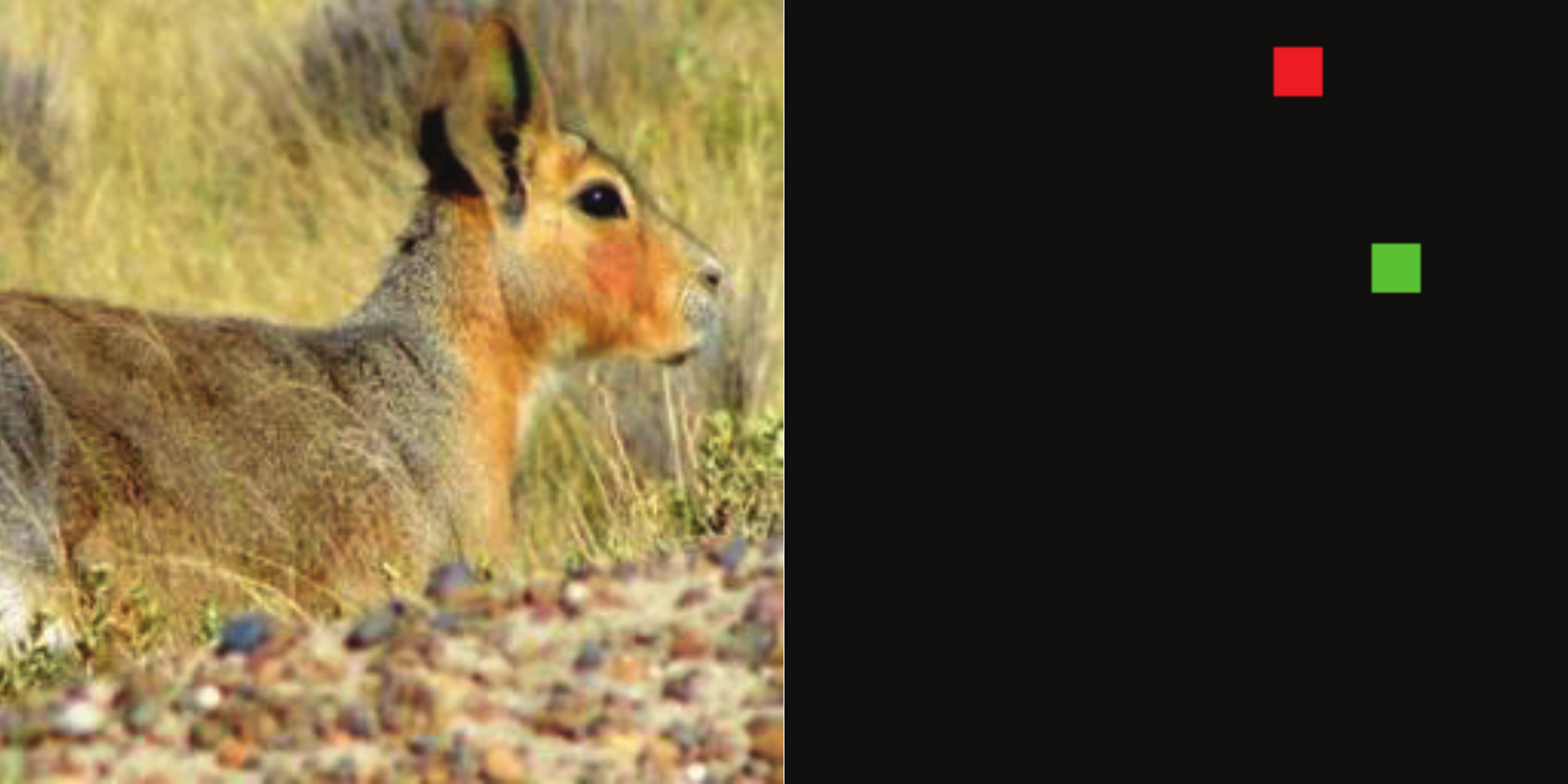}
  \end{minipage}}%
  \hspace{-1.3mm}
\subfigure[BP-saliency map]{
  \begin{minipage}[b]{0.12\textwidth}
    \includegraphics[width=\textwidth]{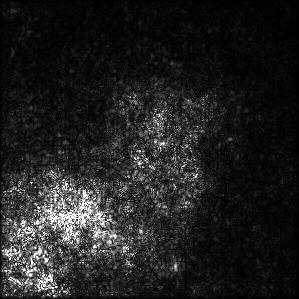}
    \includegraphics[width=\textwidth]{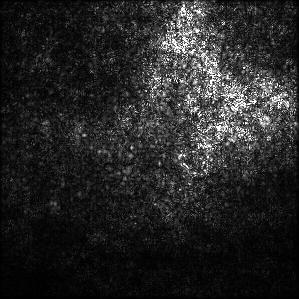}
  \end{minipage}}%
  % \vspace{-1\baselineskip}
\caption{Examples generated by Saliency Attack with different initial block sizes.}\label{fig9}
\end{figure}

We also show some examples in Figure~\ref{fig9}. It can be observed that as $k_{int}$ decreases, the generated perturbations become smaller but more salient. For instance, in the first row, the perturbation with $k=128$ roughly covers the region of the dog face, which is also the brightest region in the BP-saliency map. While the perturbation with $k=32$ or $k=16$ focuses on smaller region of the dog ear. This indicates that our Saliency Attack could find smaller and more important perturbations progressively as the $k_{int}$ decreases, which coincides with our assumption before. The perturbations are also interpretable to some extent, like perturbing the specific regions of a dog's ears, eyes or nose.

\subsection{Attacking Detection-based Defense}
To further validate the imperceptibility of AEs, we attack against two detection-based defenses that attempt to detect AEs from clean images. The first is Feature Squeezing\footnote{Implementation of Feature Squeezing: https://github.com/mzweilin/EvadeML-Zoo} \cite{42}. Its basic idea is to squeeze out unnecessary input features that may be utilized by an adversary to generate AEs. Comparing the model's prediction on the original image with its prediction on the image after squeezing, the input is likely to be adversarial if the original and squeezed inputs produce substantially different outputs. Feature Squeezing adopts color bits reduction of each pixel and spatial smoothing (local and non-local smoothing) as squeezers. We use the recommended parameters ($5-bit, 2\times2, 11-3-4$) for ImageNet dataset. Another is a CNN-based binary classifier based on a Steganography detector \cite{52}, which is designed for detecting small noise specifically and outperforms classic classification models such as Inception-v3 or ResNet. We finetune the pretrained model\footnote{https://github.com/brijeshiitg/Pytorch-implementation-of-SRNet} with AEs generated by different attacks. Concretely, we randomly select 1,000 clean images and generate the corresponding AEs with Saliency Attack and six baselines, respectively. Thus, we use $2\times7,000$ pairs of clean images and AEs as the training set to finetune the model. The training is run for 100 epochs, and the learning rate is 0.001. We test another 1,000 samples and record the detection rate (DR), which is the lower the better.

\begin{table}[h]
% \setlength{\abovecaptionskip}{-8pt}
  % \small
  \caption{Results for attacking against detection-based defenses} \label{tab4}
  % \vspace{-8pt}
  \centering
  \begin{tabular}{c|cccc}
  	\toprule
  	{Method} & {$\rm SR$} & {$\rm SR_{true}$} & {Feature Squeezing DR} & {Classifier DR} \\
  	\midrule
    Boundary & 100.0\% & 30.2\% & 64.5\% & 68.0\% \\
  	TVDBA & 96.9\% & 23.2\% & 26.5\% & 61.9\% \\
  	Parsimonious & 98.2\% & 14.1\% & 33.3\% & 89.7\% \\
    Parsimonious-sal & 96.7\% & 12.0\% & 29.7\% & 67.6\% \\
  	Square   & 99.7\% & 1.9\% & 47.8\% & 85.4\% \\
    Square-sal & 97.1\% & 22.5\% & 26.1\% & 63.1\% \\
  	Saliency (ours) & 93.6\% & 86.2\% & \textbf{21.2\%} & \textbf{50.4\%} \\
  	\bottomrule
  \end{tabular}
  % \vspace{-5pt}
\end{table}

From Table~\ref{tab4}, we can find our Saliency attack can achieve lower detection rate against both Feature Squeezing and binary classifier detection than other baselines. For Feature Squeezing, although the area of perturbation generated by Saliency Attack is quite small and thus suffering a higher risk of being denoised by squeezers, the detection rate of Saliency Attack is still better than other attacks due to our effective perturbation. For the binary classifier detector, our Saliency Attack obtains nearly $50\%$ accuracy/detection rate, which means the perturbation is invisible enough that the classifier will be difficult to converge and results in random guess. Therefore, our Saliency Attack is able to generate imperceptible AEs that evade different detection-based defenses.

\section{Conclusion}\label{sec:5}
In this paper, we propose restricting black-box attacks to perturb in salient regions to improve the imperceptibility of AEs, and propose a novel black-box attack, Saliency Attack. Experiments show that SOTA black-box attacks restricted in salient regions can indeed achieve better imperceptibility performance, while Saliency Attack further enhances this by recursively refining perturbations in salient regions. Its perturbation is much smaller, imperceptible and interpretable to some extent. Besides, we also find that the salient regions of some examples are indeed progressive with respect to their impact on model output, which is consistent with our assumption from the previous visualization studies. Finally, the evaluation demonstrates that the AEs generated by Saliency Attack are harder to be detected, which validates its imperceptibility from the defensive perspective.

Although our \textit{Refine} search provides significant benefit to the imperceptibility, it inevitably consumes more queries compared with the black-box attacks with global perturbations.
In the future, we will try to find better ways to balance the three objectives of query efficiency, imperceptibility and attack success rate for black-box attacks. Furthermore, we will apply the salient region to more black-box attacks to improve their imperceptibility. Our Saliency Attack could also be tested against other kinds of defenses, such as robustness-based defenses (adversarial training, network distillation, etc.), to validate its effectiveness.

% \begin{acks}
% \end{acks}

\bibliographystyle{ACM-Reference-Format}
\bibliography{sample-base}

\appendix

\section{Appendix}

\subsection{MAD Metric}\label{appendix:sec:MAD}
Most Apparent distortion (MAD) metric \cite{39} is one of the state-of-the-art full-reference image quality assessment methods. MAD attempts to merge two separate strategies for two kinds of distorted images, respectively. For high-quality images with near-threshold distortion (just visible), MAD focuses on detection-based strategy to look for distortions in the presence of the image. While for low-quality images with suprathreshold distortion (clearly visible), MAD focuses on appearance-based strategy to look for image content in the presence of the distortions. MAD will control the weight of two strategies according to the type of distorted images. The calculation process of MAD can be summarized as following steps and we recommend interested readers to read the original literature.
\begin{enumerate}
\item Compute locations of visible distortions based on luminance images.
\item Combine the visibility map with local error image.
\item Decompose both the distorted and original images into log-Gabor subbands.
\item Calculate different statistics of each subband.
\item Calculate the adaptive blending score.
\end{enumerate}

\subsection{PFA Network for Saliency Detection}\label{appendix:sec:PFA}
Pyramid Feature Attention (PFA) network \cite{23} is a novel salient object detection method considering the different characteristics level features. Specifically, the saliency maps from low-level features contain many noises, while the saliency maps from high-level features only get an approximate area. Therefore, PFA network first devises a context-aware pyramid feature extraction (CPFE) module to get multi-scale multi-receptive-field high-level features, and then uses channel-wise attention (CA) to select appropriate scale and receptive-field for generating saliency regions. On the other hand, to refine the boundaries of saliency regions, PFA network uses spatial attention to better focus on the effective low-level features, and obtain clear saliency boundaries. After the processing of different attention mechanisms, the high-level features and low-level features are complementary-aware and suitable to generate saliency map. Besides, an edge preservation loss is proposed to guide the network to learn more detailed information in boundary localization. With these powerful feature extraction methods and attention mechanisms, PFA network can achieve robust and effective salient object detection, outperforming SOTA methods under different evaluation metrics.

\subsection{Threshold for MAD Metric}\label{appendix:sec:threshold}
\begin{figure}[ht]
\centering
\subfigure[MAD=50]{
  \begin{minipage}[b]{0.16\textwidth}
    \includegraphics[width=\textwidth]{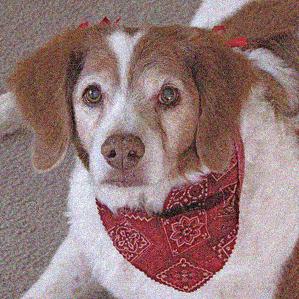}
    \includegraphics[width=\textwidth]{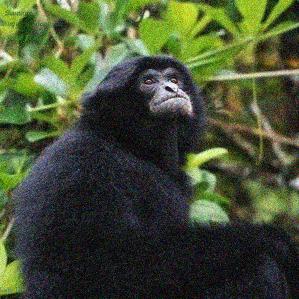}
    \includegraphics[width=\textwidth]{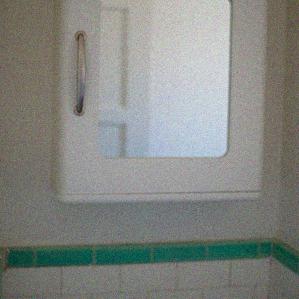}
    \includegraphics[width=\textwidth]{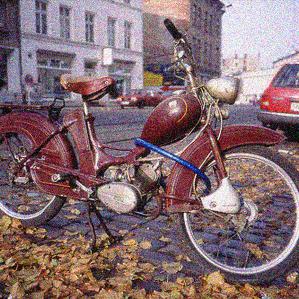}
  \end{minipage}}%
\subfigure[MAD=40]{
  \begin{minipage}[b]{0.16\textwidth}
    \includegraphics[width=\textwidth]{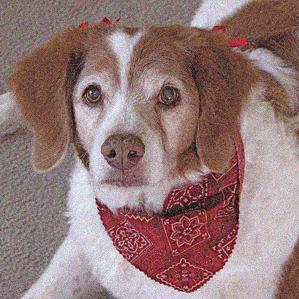}
    \includegraphics[width=\textwidth]{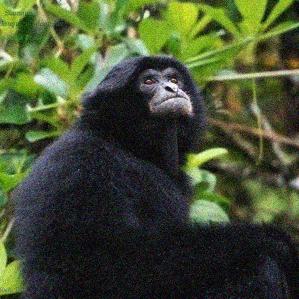}
    \includegraphics[width=\textwidth]{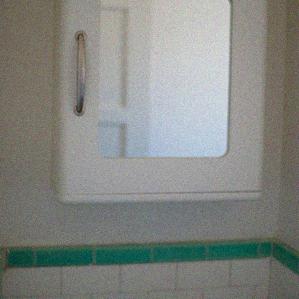}
    \includegraphics[width=\textwidth]{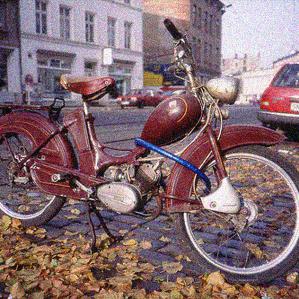}
  \end{minipage}}%
\subfigure[MAD=30]{
  \begin{minipage}[b]{0.16\textwidth}
    \includegraphics[width=\textwidth]{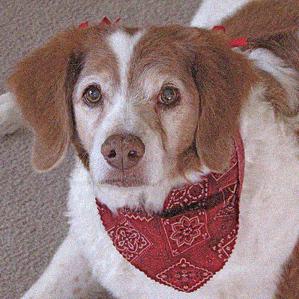}
    \includegraphics[width=\textwidth]{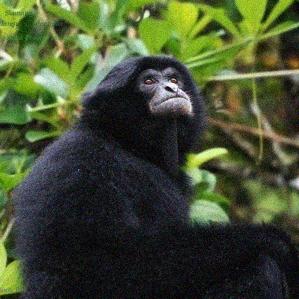}
    \includegraphics[width=\textwidth]{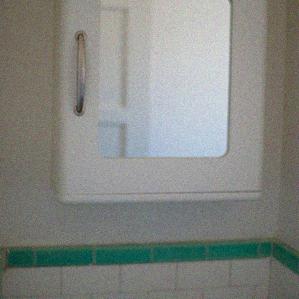}
    \includegraphics[width=\textwidth]{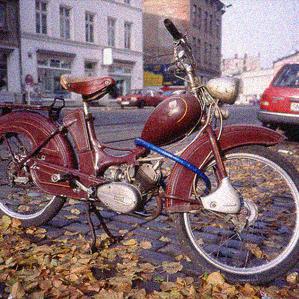}
  \end{minipage}}%
\subfigure[MAD=20]{
  \begin{minipage}[b]{0.16\textwidth}
    \includegraphics[width=\textwidth]{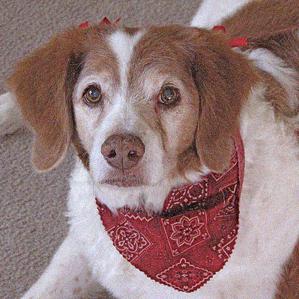}
    \includegraphics[width=\textwidth]{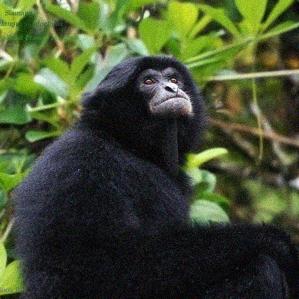}
    \includegraphics[width=\textwidth]{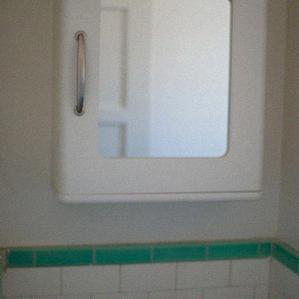}
    \includegraphics[width=\textwidth]{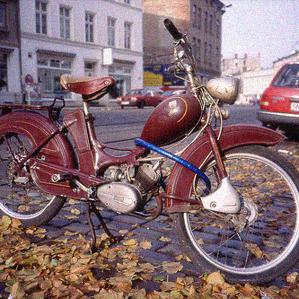}
  \end{minipage}}%
\subfigure[MAD=10]{
  \begin{minipage}[b]{0.16\textwidth}
    \includegraphics[width=\textwidth]{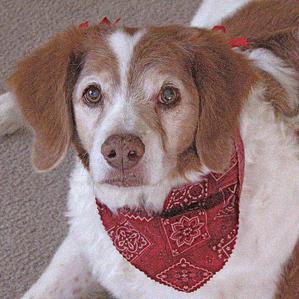}
    \includegraphics[width=\textwidth]{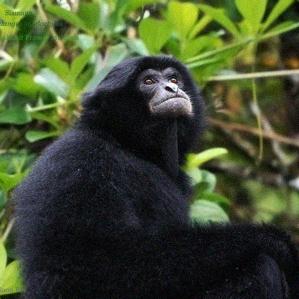}
    \includegraphics[width=\textwidth]{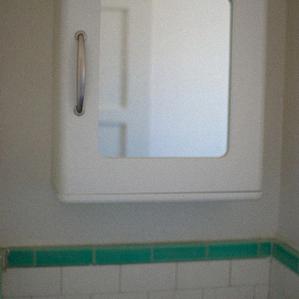}
    \includegraphics[width=\textwidth]{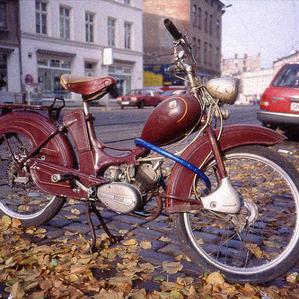}
  \end{minipage}}%
\subfigure[Original]{
  \begin{minipage}[b]{0.16\textwidth}
    \includegraphics[width=\textwidth]{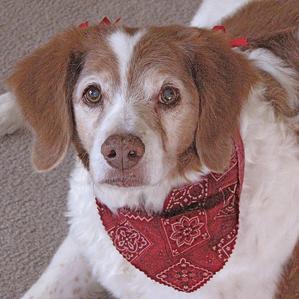}
    \includegraphics[width=\textwidth]{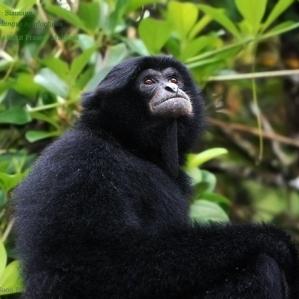}
    \includegraphics[width=\textwidth]{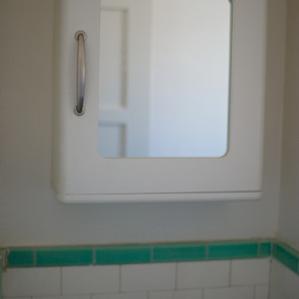}
    \includegraphics[width=\textwidth]{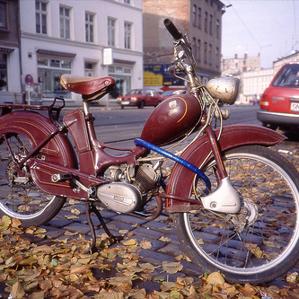}
  \end{minipage}}%
  % \vspace{-1\baselineskip}
\caption{Adversarial examples generated by boundary attack with different MAD scores. We can find the threshold $\rm MAD \le 30$ is roughly enough to indicate an imperceptible adversarial example.}\label{fig11}
\end{figure}

Since Boundary Attack \cite{45} can reduce the distortion gradually, we generate multiple adversarial examples with different MAD scores in Figure~\ref{fig11} to find a proper threshold for MAD metric.

\subsection{Scatter Plot of MAD Scores}\label{appendix:sec:scatter}
\begin{figure}[h]
\centering
\includegraphics[width=0.45\textwidth]{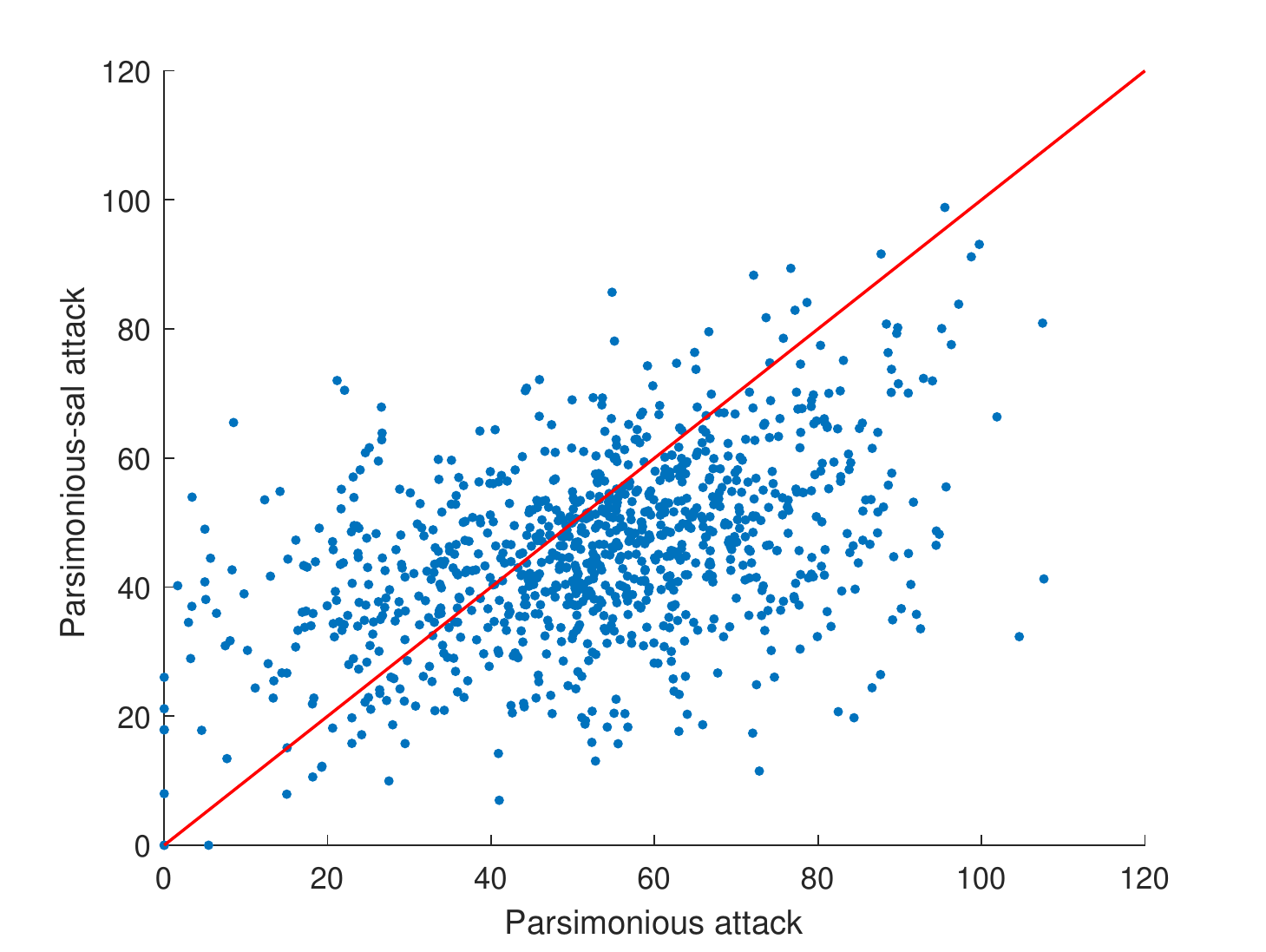}
\hfill
\includegraphics[width=0.45\textwidth]{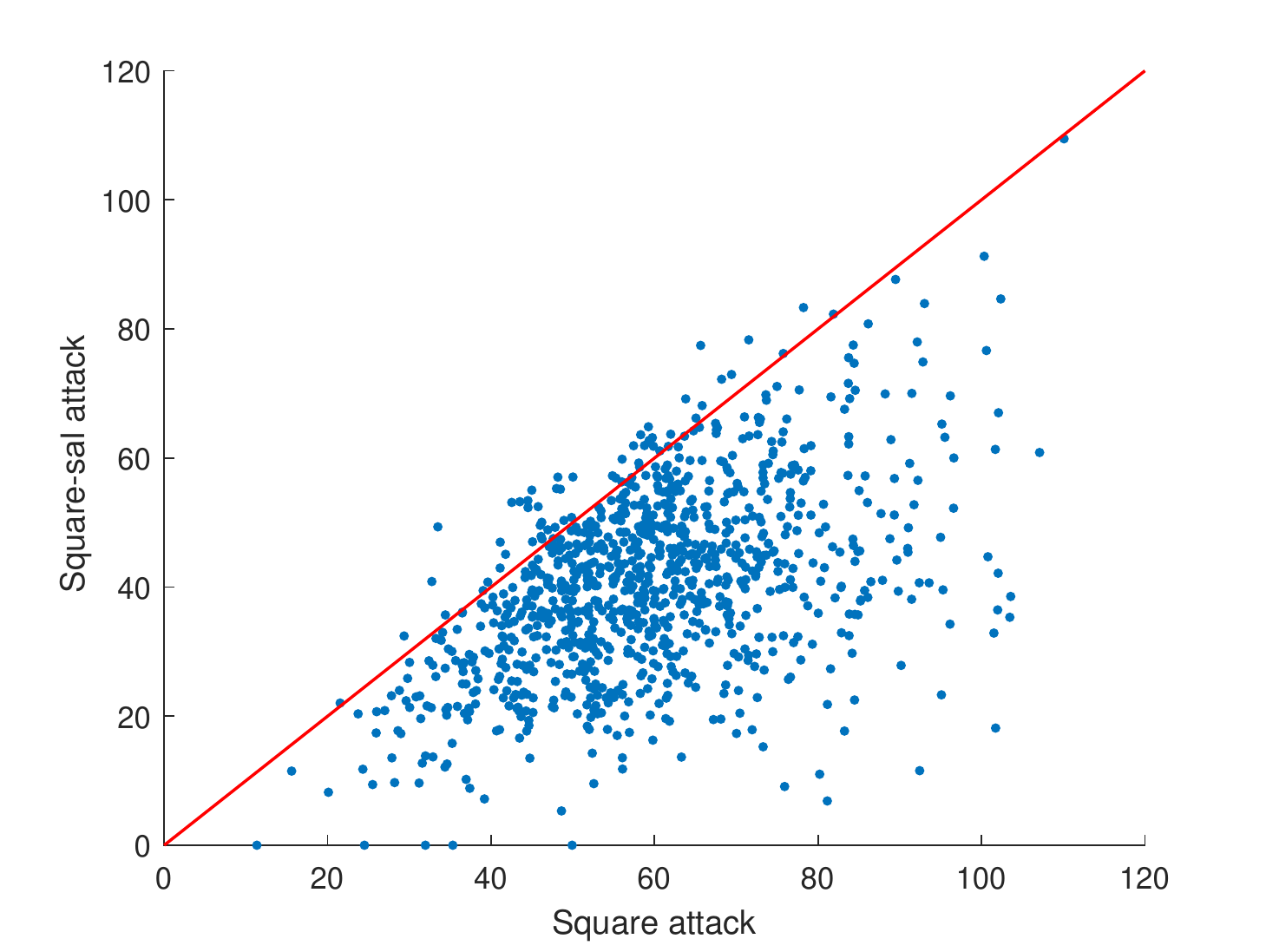}
\hfill
\caption{MAD scores of Parsimonious Attack vs. Parsimonious-sal attack and Square Attack vs. Square-sal Attack.}\label{fig12}
\end{figure}

As shown in Figure~\ref{fig12}, via restricting perturbations in salient region, 68.1\% and 94.3\% examples of Parsimonious-sal attack and Square-sal attack have better MAD scores compared with their original version, respectively. Since Parsimonious Attack uses a greedy local search while Square attack adopts a random search, it is obvious that limiting the search in salient regions is more helpful to Square Attack.

\subsection{Greedy Search in Salient Region with Different Block Sizes}\label{appendix:sec:GS}
We test different block sizes for greedy search in salient region in Table~\ref{tab5}, and choose the best one (block size equals 32) to be compared in ablation study.
\begin{table}[h!]
\caption{Greedy search in salient region with different block sizes.} \label{tab5}
\centering
\begin{tabular}{c|cccccc}
	\toprule
  {Block size} & {Avg. queries} & {$\rm SR$} & {$\rm SR_{true}$} & {$\rm L_2$} & {$\rm L_0$} & {$\rm MAD$}\\
	\midrule
  {128} & 57 & 20.6\% & 18.4\% & 7.32 & 12.8\% & 11.50\\
  {64} & 512 & 37.4\% & 31.3\% & 6.37 & 10.2\% & 15.18\\
  {32} & 2727 & 56.0\% & \textbf{50.7\%} & 4.37 & 4.7\% & 12.87 \\
  {16} & 4039 & 35.4\% & 35.2\% & 1.79 & 0.8\% & 4.84\\
	\bottomrule
\end{tabular}
% \vspace{-8pt}
\end{table}

\end{document}